\documentclass{ieeeaccess}
\usepackage{cite}
\usepackage{amsmath,amssymb,amsfonts}
\usepackage{algorithmic}
\usepackage{graphicx}
\usepackage{pifont}
\usepackage{textcomp}
\usepackage{bm}
\usepackage{url}
\usepackage{booktabs}
\usepackage{multirow}
\usepackage{multicol}
\usepackage{soul} 
\usepackage{enumitem}
\usepackage{hyperref}
\usepackage{comment}
\usepackage{lipsum}
\usepackage{enumitem}
\usepackage{makecell}
\usepackage{array}
\usepackage{ragged2e}

\makeatletter
\AtBeginDocument{\DeclareMathVersion{bold}
\SetSymbolFont{operators}{bold}{T1}{times}{b}{n}
\SetSymbolFont{NewLetters}{bold}{T1}{times}{b}{it}
\SetMathAlphabet{\mathrm}{bold}{T1}{times}{b}{n}
\SetMathAlphabet{\mathit}{bold}{T1}{times}{b}{it}
\SetMathAlphabet{\mathbf}{bold}{T1}{times}{b}{n}
\SetMathAlphabet{\mathtt}{bold}{OT1}{pcr}{b}{n}
\SetSymbolFont{symbols}{bold}{OMS}{cmsy}{b}{n}
\renewcommand\boldmath{\@nomath\boldmath\mathversion{bold}}}
\makeatother

\def\BibTeX{{\rm B\kern-.05em{\sc i\kern-.025em b}\kern-.08em
    T\kern-.1667em\lower.7ex\hbox{E}\kern-.125emX}}

\begin{document}
\history{Date of publication xxxx 00, 0000, date of current version xxxx 00, 0000.}
\doi{10.1109/ACCESS.2024.0429000}

\title{Data-driven Day Ahead Market Prices Forecasting: A Focus on Short Training Set Windows}

\author{
    \uppercase{Vasilis Michalakopoulos}\authorrefmark{1}, 
    \uppercase{Christoforos Menos-Aikateriniadis}\authorrefmark{1}, 
    \uppercase{Elissaios Sarmas}\authorrefmark{1}, 
    \uppercase{Antonis Zakynthinos}\authorrefmark{1}, 
    \uppercase{Pavlos S. Georgilakis}\authorrefmark{1}, \IEEEmembership{Senior Member, IEEE}, 
    \uppercase{Dimitris Askounis}\authorrefmark{1}
}

\address[1]{School of Electrical and Computer Engineering, National Technical University of Athens (NTUA), Athens 15780 Greece (e-mail: christoforosmenos@mail.ntua.gr)}

\tfootnote{
The work presented is based on research conducted within the framework of the Horizon Europe European Commission project CRETE VALLEY (Grant Agreement No. 101136139). The content of the paper is the sole responsibility of its authors and does not necessary reflect the views of the EC.}

\markboth
{Author \headeretal: Preparation of Papers for IEEE TRANSACTIONS and JOURNALS}
{Author \headeretal: Preparation of Papers for IEEE TRANSACTIONS and JOURNALS}

\corresp{Corresponding author: Vasilis Michalakopoulos (e-mail: vmichalakopoulos@epu.ntua.gr).}

\begin{abstract}

This study investigates the performance of machine learning models in forecasting electricity Day-Ahead Market (DAM) prices using short historical training windows, with a focus on detecting seasonal trends and price spikes. We evaluate four models, namely LSTM with Feed Forward Error Correction (FFEC), XGBoost, LightGBM, and CatBoost, across three European energy markets (Greece, Belgium, Ireland) using feature sets derived from ENTSO-E forecast data. Training window lengths range from 7 to 90 days, allowing assessment of model adaptability under constrained data availability. Results indicate that LightGBM consistently achieves the highest forecasting accuracy and robustness, particularly with 45–60 day training windows, which balance temporal relevance and learning depth. Furthermore, LightGBM demonstrates superior detection of seasonal effects and peak price events compared to LSTM and other boosting models. These findings suggest that short-window training approaches, combined with boosting methods, can effectively support DAM forecasting in volatile, data-scarce environments.

\end{abstract}


\titlepgskip=-21pt

\maketitle

\section{Nomenclature}
\subsection{Acronyms \& Abbreviations}
\begin{tabbing}
    \hspace{1.5cm} \= \kill 
    AI \> Artificial Intelligence \\
    ANN \> Artificial Neural Network \\
    ARIMA \> AutoRegressive Integrated Moving Average \\
    BM \> Balancing Market \\
    CNN \> Convolutional Neural Network \\
    DAM \> Day-Ahead Market \\
    DL \> Deep Learning \\
    DNN \> Deep Neural Network \\
    FM \> Forward Market \\
    FFEC \> Feed-Forward Error Correction \\
    FSI \> Forecast Skill Index \\
    GRU \> Gated Recurrent Unit \\
    IDM \> Intra-day Market \\
    LSTM \> Long Short-Term Memory \\
    MAE \> Mean Absolute Error \\
    MAPE \> Mean Absolute Percentage Error \\
    ML \> Machine Learning \\
    MSE \> Mean Squared Error \\
    PV \> Photovoltaic \\
    RNN \> Recurrent Neural Network \\
    RMSE \> Root Mean Squared Error \\
    SVR \> Support Vector Regression \\
    TFT \> Temporal Fusion Transformer \\
\end{tabbing}

\section{Introduction}
\label{sec:intro}
\subsection{Motivation}



\PARstart{A}{ccurate} day-ahead electricity price forecasting is a strategic necessity in liberalized power markets. Market participants are increasingly exposed to financial risks driven by the rising volatility of electricity wholesale prices—volatility that stems not only from demand fluctuations and generation intermittency but also from evolving regulatory frameworks, bidding rules, and cross-border interconnections \cite{maciver2021electrical}. In this environment, the ability to anticipate price spikes, imbalance costs, or low-margin conditions with even marginally improved accuracy can translate directly into competitive advantage or avoided financial loss \cite{manfre2023hybrid}.

The European electricity market landscape, shaped by the Clean Energy Package and further harmonized through the EU Target Model, has shifted toward real-time responsiveness and market coupling. As trading horizons shrink and operational constraints tighten, forecasting tools must now support continuous adaptation. Traditional models based on extensive historical data—often spanning multiple years—fail to reflect recent structural shifts or evolving participant behavior. In fast-changing regulatory contexts or new market entries (e.g., asset commissioning, portfolio restructuring), historical depth is not a luxury forecasters can rely on.

Price formation in Day-ahead Markets (DAMs) is now influenced by a combination of complex drivers: dynamic merit order reshuffling due to renewable and gas pricing \cite{shimomura2024beyond}, shifting congestion patterns \cite{kazemtarghi2024dynamic}, balancing reserve valuations, and increasingly, by cross-zonal flows managed through EUPHEMIA-based market coupling \cite{ovaere2023effect}. Accurate modeling under such conditions requires not just predictive precision but responsiveness to short-term signals. Short training windows offer a pragmatic solution—especially when system behavior is non-stationary or when forecasts must be deployed with minimal delay or localized data only.

Moreover, recent developments in grid flexibility \cite{michalakopoulos2024machine}, prosumer participation \cite{toderean2024demand}, and decentralized optimization require forecasting approaches that are not only accurate in average metrics but also sensitive to extremes—such as peak price spikes or demand troughs \cite{silva2022market}. These are the events that test the limits of asset profitability and system resilience. In this context, models that learn quickly from short, recent histories and adapt to seasonal dynamics are increasingly valuable. Against this backdrop, our work focuses on a critical but underexplored question: how can machine learning (ML) models be optimized to deliver accurate price forecasts when constrained to short historical datasets? This question is directly relevant to energy traders, virtual power plant operators and Transmission System Operators (TSOs) working in highly dynamic or data-scarce environments \cite{tschora2022electricity}.

We propose a focused benchmark study across three European markets—Greece, Belgium, and Ireland—selected for their distinct regulatory, climatic, and operational profiles. Our modeling pipeline relies exclusively on forecasted features, replicating the real information set available to market actors at the time of DAM bidding. This ensures the validity of results for operational use.

\subsection{Related Work}

DAM price forecasting has been a highly active area of research over the past decade that has increasingly drawn the attention of academic communities while many state-of-the-art (SOTA) practices have been proposed and implemented, paving the way for accurate and reliable forecasting models. According to the comprehensive review of the methods proposed for market price forecasting spanning 2014 to 2021 by Lago et al. \cite{LIT_1}, the field is highly diverse with solutions belonging to three categories, i.e. statistical, ML and hybrid methods. For the scope of this study, the focus will be on ML-oriented studies published after 2021 to highlight the latest advances in the sector. Additionally, the reviewed studies are further categorized based on the nature of their contribution, distinguishing between those that focus on model-centric innovations, those that emphasize data-centric and scenario-based strategies and those that utilize both approaches. 

\subsubsection{Model-based approaches}
Initially focusing on research that mainly involves novelty surrounding a ML model's architecture and several methods aimed at optimizing it, Bozlak et al. \cite{LIT_5} proposed a hybrid model, by merging a Convolutional Neural Network (CNN) with a Long Short Term Memory (LSTM) model, trained on electricity pricing data from the German electricity market, incorporating total hourly consumption as an exogenous input. The dataset spanned from early 2019 until late 2021 and the proposed model generally outperformed both a simple LSTM and a Seasonal Autoregressive Integrated Moving Average model with exogenous variables (SARIMAX), although the former provided better results when tested on data from 2020 due to increased noise and volatility. Experimentation with model hybridization was also the focus of Qiang Tan et al. \cite{LIT_4} who formed 4 seasonal datasets each consisting of five years of historical data from the Australian national electricity market and trained a 3-stage hybrid model. The model consisted of an initial time-series decomposition to Intrinsic Mode Functions through improved complete ensemble empirical mode decomposition with adaptive noise followed by a CNN module for local feature extraction and a stacked sparse denoising autoencoder for robust nonlinear feature learning. The proposed solution proved to be consistently the best performing model compared to traditional ML and Deep Learning (DL) models as well as hybrid decomposed and non-decomposed variants. In the same manner, Li et al. \cite{LIT_8} used historical hourly data from a regional Chinese market to train a proposed hybrid combination of a Gated Recurrent Unit (GRU) with a Light Gradient Boosting Machine (LightGBM), improved through bayesian optimization. The training data spanned over a year and the testing period over 3 months covering multiple temporal patterns. Their method proved consistently more accurate than it's non-optimized variant and a baseline GRU model, improving forecasts by 3.3\% and 1.6\% respectively. Similarly focusing on merging hybrid models with further enhancements, Mubarak et al. \cite{LIT_2} trained a combination of a hybrid model consisting of a CNN and a Bidirectional LSTM (BiLSTM) and an auto-regression model, for non-linear and linear modeling respectively, on one year of hourly data from the UK and German electricity markets separately and experimented with different  hyperparameter optimization techniques like random search, a genetic algorithm and particle swarm optimization (PSO). The results showed that PSO achieved a decrease in Root Mean Square Error (RMSE) of 7-9\% and Mean Absolute Error (MAE) of 10-19\% compared to the other techniques, including a non-optimized one in the UK market, while the decrease was around 10-17\% with respect to RMSE and 11-24\% with respect to MAE in the German market. Also, on a comparable yet distinct approach, Kitsatoglou et al. \cite{LIT_6} trained multiple statistical, ML and DL models on hourly historical pricing, load and generation data over the span of a year from the German market and experimented with different ensemble strategies to further enhance the robustness and accuracy of day-ahead forecasts. Notably, the methods involved selected the single best-performing model from the previous day and the model with the lowest average hourly MAE over the past 20 days. Additionally, ensemble strategies were tested, such as combining the top models from the last three days using various weighting schemes, as well as taking a simple average of their predictions. Tested only on 28 days of data, the Multilayer Perceptron (MLP) was the best single model and all custom ensemble models outperformed all individual models except MLP while the strategy involving the best performing model per hour had consistently the best overall performance. Overall, these approaches by leveraging architectural innovations have demonstrated consistent performance gains over conventional methods.

\subsubsection{Data-centric or scenario-based approaches}
\label{subsec:data-centric}
Several studies focused on strategies that surround training data and certain scenarios. For example, Yorat et al. \cite{LIT_10} focused on the Turkish market and used lagged electricity prices, exchange rates, stock market indices, natural gas prices, and cross-border electricity prices from neighboring countries to construct a comprehensive feature set for training regressors like Multi-linear Regression (MLR) and ARIMA as well as an Extreme Gradient Boosting (XGBoost) model. The results showed the superiority of the latter, which outperformed both MLR and ARIMA in all evaluation metrics with a $R^2$ of 95.27\%. Likewise, Tschora et al. \cite{LIT_12} trained an auto-regressive model and several ML models like a Support vector Regressor (SVR), a Random Forest model, a Deep Neural Network (DNN) and a CNN on historical data from the German, French and Belgian markets while focusing on several scenarios and data refinements. Specifically, they experimented with two different periods of 5 years, a single-country and multi-country training scenario and with an enhanced version of the original dataset by adding features like forecasts for demand and generation, stock market indices and cross-border pricing. The results showed that DNNs and SVRs, trained with the enriched feature set, consistently outperformed traditional models and benchmarks across all markets by achieving up to 15\% lower RMSE. In a slightly different approach, Sun et al. \cite{LIT_7} used historical pricing, load, generation and weather data over 15 minute intervals for one year from the Chinese market and trained a gradient-boosting decision tree (GBDT) and enhanced it with several algorithms. Specifically, in the first experiment, a segmentation based on load rate was performed by comparing price and system load rate curves followed by the second experiment, where they also performed parameter optimization for the GBDT. In the third one, the authors added an additional final deviation correction step in peak and flat time periods to calibrate the predictions based on past deviation trends. The results demonstrated a consistent improvement in forecast accuracy across these three experimental schemes, culminating in a Mean Absolute Percentage Error (MAPE) reduction of approximately 50-60\% from the first to the third approach. In summary, these studies highlight the critical role of enriched feature sets and carefully designed training scenarios in improving forecast accuracy.

\subsubsection{Mixed approaches}
Some researchers sought to integrate both model-centric and data-centric or scenario-based strategies within a unified forecasting framework. Such an example is the study of Micu et al. \cite{LIT_11} who trained a hybrid model, combining a CNN with BiLSTMs, on a dataset formed from 3 years of data in the United Kingdom market. The input features included historical pricing, demand, generation, weather and calendar data while the results showed promising forecast accuracy with MAPE as low as 3.3\% for days with generally stable prices, but the model struggled with high volatility and large price spikes. In a related approach, Pavicevic et al. \cite{LIT_9} formed a dataset from hourly historical pricing, weather and time data spanning over a period of nine years from the Hungarian Day-ahead spot market and used it to train a fully connected neural network (Dense), CNN, LSTM, auto-regressive LSTM and two hybrid models combining a Dense with an LSTM and a CNN, respectively. Additionally, they experimented with an expanded dataset from 2011 onward and improved it by adding a seasonal difference feature capturing the difference in pricing per hour compared to the same hour one year before. The results indicated a superior performance from the hybrid model containing convolutional layers over both datasets, while it benefits from the expansion providing more robust forecasts. Lastly, focusing on more complex models and data-based scenarios, Nitsch et al. \cite{LIT_3} utilized the AMIRIS simulator to generate more than 100 total scenarios each containing a full year of hourly data based on different combinations of flexibility and renewable energy capacity. The authors conducted experiments by separately training an N-Beats model and a Temporal Fusion Transformer (TFT) in multiple training / testing splits by varying the number of scenarios used for training and testing. As a result the TFT offers consistently improved performance than N-Beats and other naive approaches mainly because of it's ability to use future co-variates while increased flexibility and a higher number of training scenarios led to significantly better performance across all models. Ultimately, by integrating all these different strategies together, mixed approaches aim to achieve an even more robust generalization and adaptability across diverse market conditions.

\subsection{Contribution}

This study adopts an approach aligned with the second category (Section \ref{subsec:data-centric}), since it uses data enhancements and specific training scenarios that benchmark multiple ML models. Compared to the reviewed literature, the variable look-back windows used by the proposed training pipeline and the analysis per calendar month and season are the determining factors that demonstrate the innovative contributions to the field. Specifically, the state-of-the-art practices used in this paper based on recent literature as well as the novel advancements proposed can be summarized as follows:

\begin{itemize}
    \item \textbf{Short-History ML Forecasting:} Demonstrates the effectiveness of ML models trained on short historical DAM prices and energy data, addressing the challenge of non-stationary system behavior and the limited relevance of older data in fast-evolving electricity markets.
    
    \item \textbf{Realistic and Operational Benchmarking:} Establishes a practical benchmarking framework across three diverse European markets (Greece, Belgium, Ireland), using only forecasted features available at the time of bidding to simulate real-world decision-making.
    
    \item \textbf{Focus on Extreme and Seasonal Events:} Emphasizes model sensitivity not only to average performance but also to price extremes (e.g., spikes, troughs), reflecting the real financial and operational risks faced by market participants for each season.
    
    \item \textbf{Evaluation of ML Models with a Focus on Boosting Trees}: Includes a diverse set of forecasting models, including LightGBM, CatBoost, XGBoost, LSTM, and a naive baseline. The superior performance of boosting tree models in short-history, high-variability DAM scenarios is demonstrated.

\end{itemize}

\subsection{Structure}

The remainder of this paper is organized as follows. Section \ref{section:methodology} outlines the methodology of our experiments, including data pre-processing steps, forecasting methods, and a brief overview of European market structures, with a focus on the Greek market. Section \ref{sec:evaluation} presents the benchmark analysis across three market examples. Section \ref{sec:conclusions} provides concluding remarks and discusses future directions.

\section{Methodology} \label{section:methodology}
\begin{figure}[!t]
    \centering
    \includegraphics[width = \columnwidth]{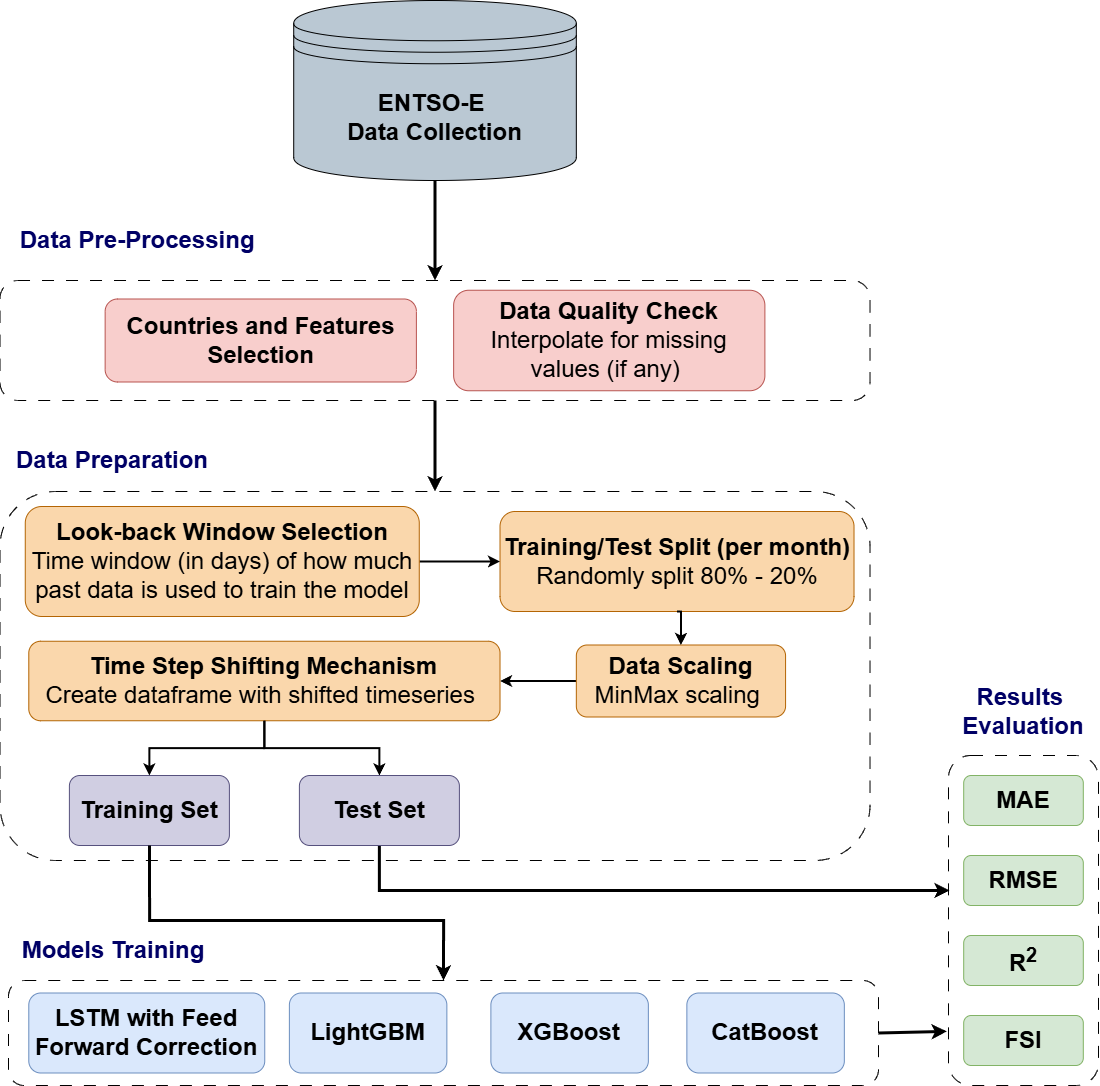}
    \caption{{Proposed methodology for Day-Ahead Market prices forecasting under different look-back windows and monthly training/test set split}}
    \label{fig:methodology}
\end{figure}

\subsection{Data Pre-Processing and Preparation}
The modeling methodology and its main analytical stages are illustrated in Fig. \ref{fig:methodology}. Initially, data was collected using the European Network of TSOs for Electricity (ENTSO-E) Transparency Platform \cite{ENTSOE}. Historical DAM prices for the electricity markets of Greece, Belgium, and Ireland were collected for the 2023 calendar year. Historical forecast data for electricity demand, RES generation, total generation and transmission net flows were also collected to serve as additional features aiming at improving DAM prices prediction accuracy. Linear interpolation was applied in case of missing values in both DAM prices and additional features. The decision to train the models using historical forecasted features corresponding to the DAM price prediction period, rather than actual values from the previous day, is motivated by two key factors: they reflect the system operators' most up-to-date view of market conditions, and they are available at the time when market participants need to make their DAM price predictions. 

The modeling pipeline differs from most related works due to the selection of various look-back windows and the decision to train and test models independently for each month of the year. More specifically, in this work a thorough investigation of different look-back windows is performed, covering 7, 14, 30, 45, 60 and 90 days. These look-back windows define the amount of past data (in days) used to train and test the models. An 80\% training and 20\% test set split ratio applies to each month of the dataset, ensuring that enough historical data is available to match the selected look-back window. As a result, the earliest training date that can be randomly selected within a month depends on the window size, i.e. for a 45-day window, the earliest day used for training would be 15/02/2023. In the 45-day look-back window case, and for longest look-back windows (e.g. 60 or 90 days) predictions for January-March cannot be conducted depending on the number of past days needed for training.


Data normalization and time-step shifting are essential preprocessing steps before feeding the data into LSTM, LightGBM, XGBoost and CatBoost models. To ensure consistency across features, MinMax scaling is applied independently to the training and test datasets, based on the transformation shown in Eq. \eqref{MinMax:std} and Eq. \eqref{MinMax:scaled}: 
\vspace{-2mm}

\begin{equation}\label{MinMax:std}
X_{\text{norm}} = \frac{{X - \min(X)}}{{\max(X) - \min(X)}}, \hspace{1cm} \forall \hspace{0.1cm} X \in \chi
\end{equation}

\begin{figure}[!t]
    \centering
    \includegraphics[width = \columnwidth]{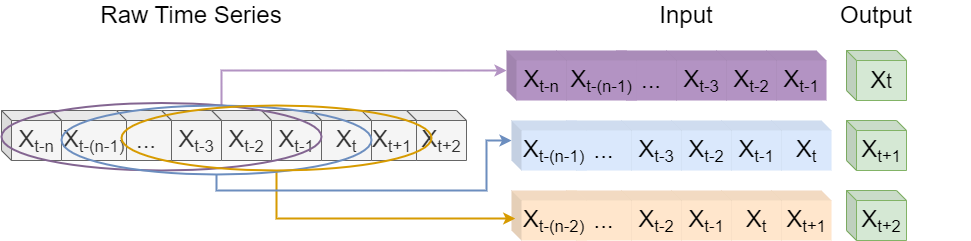}
    \caption{{Time Step Shifting Mechanism for training/test timeseries creation}}
    \label{fig:data_reshape}
\end{figure}

\vspace{-2mm}

\begin{equation}\label{MinMax:scaled}
X_{\text{scaled}} = X_{\text{norm}} \cdot (L_{up} - L_{low}) + L_{low},  \hspace{1cm} \forall \hspace{0.1cm} X \in \chi
\end{equation}

where $\chi$ represents the set of input features $X$ and $\{L_{low},L_{up}\}$ are the lower and upper boundaries of the desired feature range. Fig. \ref{fig:data_reshape} shows the time-step shifting mechanism applied in this work to create timeseries of historical data, where $n=24$ stands for the previous 24 hourly time steps. 

\subsection{Day-Ahead Market Prices Forecasting Methods}

The methodologies applied for the efficient execution of the experiment include several boosting models and a RNN. 

\subsubsection{Boosting models}

Boosting is a ML ensemble technique that creates several weak models sequentially, typically decision trees, trained to minimize the residual errors of the preceding models. The model is updated at each time step $t$ leading to a stronger model with improved accuracy based on the formula shown in Eq. \eqref{Boosting}:

\begin{equation}\label{Boosting}
f_{t}(x) = f_{t-1}(x)+ \alpha \cdot l_t(x), \hspace{0.5cm} \alpha \in (0,1]
\end{equation}
where $f_t(x)$ is the updated model at a time step \textit{t}, $l_t(x)$ is the weak learner trained on the residual error of $f_{t-1}(x)$ at time step \textit{t} while $\alpha$ is the learning rate. The Gradient Boosting is an extension to this technique and is also based on Eq. \eqref{Boosting} but instead of training each new weak learner on the residual errors of the previous model, the training occurs on the pseudo residual errors. These are simply the negative gradients of a chosen differentiable loss function per sample and is shown in Eq. \eqref{Grad_Boosting}:
\begin{equation}\label{Grad_Boosting}
r_i^{(t)} = - \left[ \frac{\partial{L(y_i,\hat{y_i})}}{\partial{\hat{y_i}}} \right]_{\hat{y_i}=f_{t-1}(x_i)}
\end{equation}
where $r_i^{(t)}$ is the pseudo-residual of the $i^{th}$ sample at the time step $t$, $L(y_i,\hat{y_i})$ is the differentiable loss function, $y_i$ is the true target value for the $i^{th}$ instance and $\hat{y_i}$ is the predicted value for the $i^{th}$ instance.

\paragraph{XGBoost model}

XGBoost, which stands for Extreme Gradient Boosting, shares the same principles of gradient boosting shown in Eq. \eqref{Boosting} and Eq. \eqref{Grad_Boosting} and performs level-wise split of the nodes. Also instead of training each weak learner on the pseudo-residual errors of the last model, it adds the weak learner that minimizes a regularized objective function $\mathcal{L}^{(t)}$ at each iteration shown in Eq. \eqref{XG_obj} which incorporates both first and second-order derivatives of the loss function and a regularization term $\Omega (l_t)$ which penalizes the complexity of the chosen weak learner as shown in Eq. \eqref{XG_reg} :
\begin{equation}\label{XG_obj}
\begin{gathered}
\mathcal{L}^{(t)} = \sum_{i=1}^{n} L(y_i,\hat{y_i}^{(t)}) + \Omega(l_t) \\
L(y_i,\hat{y_i}^{(t)}) \approx L(y_i,\hat{y_i}^{(t-1)}) + g_i l_t(x_i) + \frac{1}{2} h_i l_t^2(x_i)
\end{gathered}
\end{equation}
\begin{equation}\label{XG_reg}
\Omega(l_t) = \gamma T  + \frac{1}{2} \lambda \sum{w_j^2}
\end{equation}
where $g_i$ and $h_i$ are the gradient and hessian, respectively, of the last model's loss function, $T$ is the number of leaves the weak learner has, $\lambda$ is the L2 regularization term which helps avoid overfitting, $\gamma$ is the weak learner complexity penalty and $w_j$ is the weight of leaf \textit{j}. In this way, the XGBoost model considers both the prediction accuracy and the model complexity for each iteration, offering more accurate and robust predictions.

\paragraph{CatBoost model}

With traditional gradient boosting, each weak learner is trained on residual errors from the same data the model trained on which causes potential data leakage while categorical features must be encoded manually in order to be properly processed. Categorical Boosting (Catboost) aims to solve these problems by introducing ordered boosting and leak-free target-based encoding while splitting each node on the same feature with the same threshold leading to symmetric trees. Specifically, instead of training each weak learner on the residual errors shown in Eq. \eqref{Grad_Boosting}, it trains it on ordered pseudo-residual errors shown in Eq. \eqref{Cat_res} where each residual per instance is computed based on the instances that preceded it in a given permutation $\sigma$.
\begin{equation}\label{Cat_res}
r_i^{(t)} = -\left[\partial \frac{L\left( y_i, f_{t-1}^{\sigma(i)-1}(x_i)\right)}{\partial \hat y_i}  \right]
\end{equation}
where $\sigma(i)$ is the position of the instance $i$ in the permutation and $f_{t-1}^{\sigma(i)-1}(x_i)$ is the prediction of the previous model trained on the first $\sigma (i)-1$ instances of the permutation. In addition, regarding categorical encoding, the CatBoost model replaces each instance of a categorical feature with the mean target value of the same category of preceding instances in a given permutation $\sigma$, as shown in Eq. \eqref{Cat_enc}.
\begin{equation}\label{Cat_enc}
\hat{x}_k^i = \frac{\sum\limits_{x_j \in \mathcal{D}_k} \mathbf{1} _{\{x_j^i = x_k^i\}} \cdot y_j + a p}{\sum\limits_{x_j \in \mathcal{D}_k} \mathbf{1} _{\{x_j^i = x_k^i\}} + \alpha}
\end{equation}
where $\hat{x}_k^i$ is the encoding of the categorical feature $i$ of instance $k$, $\mathcal{D}_k = \left\{ x_j : \sigma (j) < \sigma (k) \right\}$ the set of instances before the instance $k$, $p$ is the prior target value and $\alpha$ is a regularization constant controlling the influence of the prior target value. In summary, it extends the gradient boosting framework by effectively eliminating data leakage, improving generalization and facilitating categorical feature encoding.

\paragraph{LightGBM model}
The Light Gradient Boosting Machine (LightGBM) follows the basic principles of gradient boosting and offers distinct advantages compared to the previous models. Initially, it optimizes the data through a pre-processing step that merges mutually exclusive features into a single feature to reduce dimensionality and performs leaf-wise node splitting based on the leaf with the highest loss reduction (gain). For the latter, it discretizes the continuous features into bins and each one stores the sum of gradients and hessians of the samples that fall into it while the total calculation of the gain in a specific split point is shown in Eq. \eqref{GBM_gain}:

{\footnotesize
\begin{equation}\label{GBM_gain}
\begin{gathered}
Gain(s) = \frac{G_{L(s)}^2}{H_{L(s)}+\lambda} + \frac{G_{R(s)}^2}{H_{R(s)}+\lambda} - \frac{(G_{L(s)}+G_{R(s)})^2}{H_{L(s)}+H_{R(s)}+\lambda} \\
G_{L(s)} = \sum_{b \le s} G_b \hspace{0.1cm},\hspace{0.1cm} H_{L(s)} = \sum_{b \le s} H_b \\
G_{R(s)} = \sum_{b \ge s} G_b \hspace{0.1cm},\hspace{0.1cm} H_{R(s)} = \sum_{b \ge s} H_b,
\end{gathered}
\end{equation}
}
where $s$ is the split point, $(G_b,H_b)$ are the sum of gradients and hessians of all samples that fall into bin $b$, $(G_L,H_L)$ are the total gradient and hessian for all the bins less than or equal to bin $s$, $(G_R,H_R)$ are the total gradient and hessian for all the bins greater than bin $s$ and $\lambda$ is the L2 regularization term on the leaf weights. Although this equation implies the use of the full dataset, this practice can be slow for large datasets and could hurt the accuracy of the model. For this reason, LightGBM performs non-uniform sampling by keeping the top \textit{a\%} of samples with the largest gradients and $b\%$ random samples from the rest and applies correction to their gradients and hessians as shown in Eq. \eqref{GBM_goss}:
\begin{equation}\label{GBM_goss}
g_i|h_i \leftarrow
\begin{cases}
g_i|h_i, & \text{if } i \in \text{large-gradient | hessian} \\
\frac{1 - a}{b} \cdot g_i|h_i, & \text{if } i \in \text{small-gradient | hessian}
\end{cases}
\end{equation}
This way the model uses a subset of data for significantly reduced computational cost with adjusted gradients and hessians to maintain unbiased estimations and increased efficiency.

\subsubsection{Recurrent Neural Networks}
\begin{figure}[h]
    \centering
    \includegraphics[width=\columnwidth]{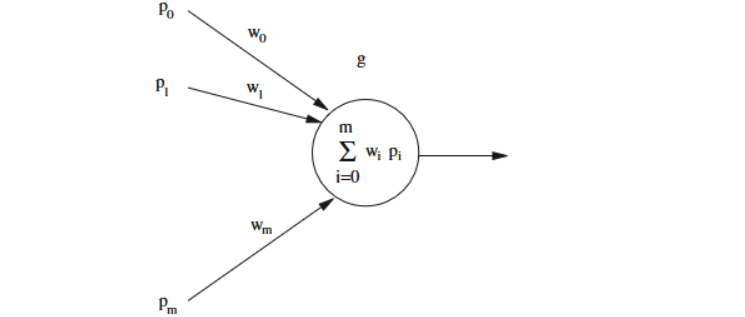}
    \caption{An artificial neuron architecture\cite{Models_2}}
    \label{fig:Perceptron}
\end{figure}

\begin{figure}[h]
    \centering
    \includegraphics[width=\columnwidth]{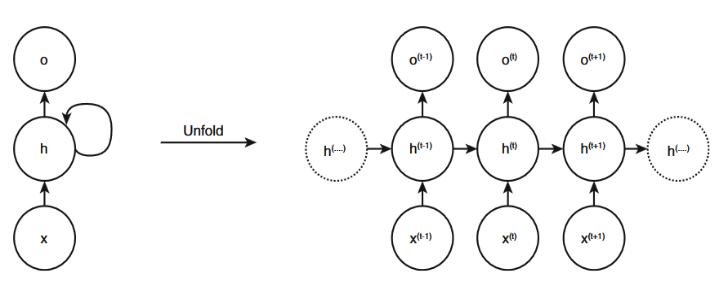}
    \caption{A simple RNN neuron (left) and its unrolled form over sequential time steps (right) \cite{Models_3}.}
    \label{fig:RNN}
\end{figure}

Aside from boosting models, artificial neural networks have transformed the forecasting industry by efficiently leveraging big data, leading to improved seasonal/trend detection and predictive accuracy. Their main difference from previous models lies in the simulation of layers of artificial neurons, connected through parameterized pathways which are learned through repetitive training, as shown in Fig. \ref{fig:Perceptron}. 
where $p_i$ is the output of neuron $i$ from the previous layer and $w_i$ is the weight of the pathway that links this neuron with neuron $i$ of the previous layer. This structure facilitates the identification of complex patterns and dependencies in the data, while several architectures extended this design, such as RNNs which have been the networks most widely used for sequence prediction problems \cite{Models_1}. Specifically, they were designed to handle sequential data and perform temporal modeling while their recurrent nature can be attributed to an additional connection that transfers the output of a neuron in a specific time step $t$ to itself as an additional input in the next time step $t+1$ called hidden state $h_t$, as shown in Fig. \ref{fig:RNN}.
This way the models can keep information from previous time steps to influence the forecasting process but they are limited only to recent past information, are prone to overfitting and suffer from the vanishing/exploding gradient problem.

\paragraph{Long Short Term Memory with Feed-Forward Error Correction}

\begin{figure}[h]
    \centering
    \includegraphics[width=\columnwidth]{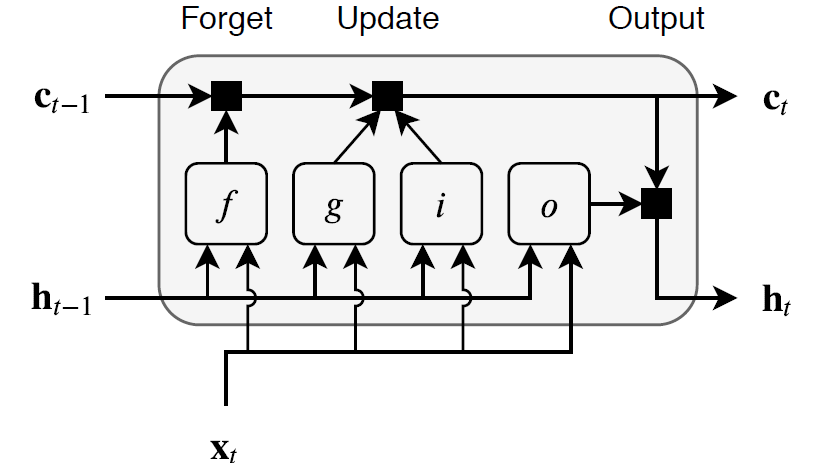}
    \caption{The architecture of an LSTM neuron \cite{Models_4}.}
    \label{fig:LSTM}
\end{figure}

To solve these problems, an LSTM model with Feed-Forward Error Correction (FFEC) was developed. The LSTM architecture, originally introduced by \cite{LSTM} and subsequently enhanced by \cite{LSTM_Gers}, is a specialized type of RNN which introduced gating mechanisms that control the flow of information within the neuron and a second transferable state in addition to the hidden state, called the cell state $c_t$, as shown in Fig. \ref{fig:LSTM}. 
The $f$ represents the Forget Gate which decides what part of the information from the previous cell state to forget as shown in Eq. \eqref{Forget_gate}, $g$ is the Candidate Cell State which proposes new information to be potentially added to the cell state, as shown in Eq. \eqref{Canditate_Cell_State}, $i$ is the Input Gate responsible for the amount of information from the candidate cell state to add to the cell state, as shown in Eq. \eqref{Input_Gate} and $o$ is the Output Gate which controls how much information from the cell state should be exposed and added to the hidden state, as shown in Eq. \eqref{Output_Gate}. Additionally, the method of update for both states is shown in Eq. \eqref{Cells_update}

\begin{equation}\label{Forget_gate}
f^{(t)} = \sigma_g(W_fx^{(t)} + R_fh^{(t-1)}+b_f)
\end{equation}
\begin{equation}\label{Canditate_Cell_State}
g^{(t)} = \sigma_c(W_gx^{(t)} + R_gh^{(t-1)}+b_g)
\end{equation}
\begin{equation}\label{Input_Gate}
i^{(t)} = \sigma_g(W_ix^{(t)} + R_ih^{(t-1)}+b_i)
\end{equation}
\begin{equation}\label{Output_Gate}
o^{(t)} = \sigma_g(W_ox^{(t)} + R_oh^{(t-1)}+b_o)
\end{equation}
\begin{equation}\label{Cells_update}
\begin{gathered}
c^{(t)} = f^{(t)} \odot c^{(t-1)} + i^{(t)} \odot g^{(t)} \\
h^{(t)} = o^{(t)} \odot \sigma(c^{(t)})
\end{gathered}
\end{equation}
where $W_i$ is the weight matrix applied to the input vector $x^{(t)}$, $R_i$ is the recurrent weight matrix applied to the hidden state $h^{(t-1)}$, $\sigma_c$ is the activation function of the state which usually is the hyperbolic tangent function and $\sigma_g$ is the activation function of the gates which usually is the sigmoid function.

\begin{figure}[!t]
    \centering
    \includegraphics[width = \columnwidth]{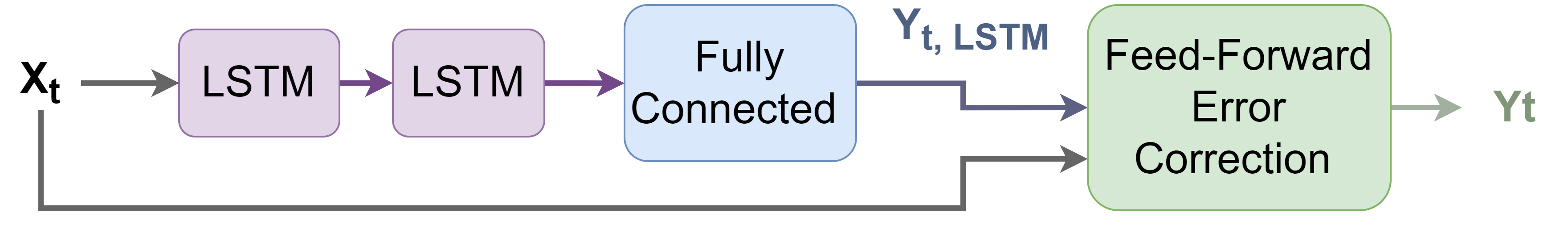}
    \caption{{LSTM with Feed-Forward Error Correction Architecture \cite{LSTM_Menos}}}
    \label{fig:LSTMmodel}
\end{figure}

In this work, the typical LSTM architecture is integrated with a Feed-Forward Error Correction mechanism, inspired by the architecture proposed in \cite{LSTM_Menos}. The developed model is illustrated in Fig.~\ref{fig:LSTMmodel}.
. Specifically, the input features \( X_t = [x_{t-n}, \ldots, x_{t-3}, x_{t-2}, x_{t-1}] \) are used to train the LSTM, which generates a preliminary prediction \( Y_{t,\text{LSTM}} \). This initial forecast is then refined using a Feed-Forward Neural Network (FFNN), which takes as input both the LSTM prediction and the historical data. The FFNN produces an improved estimate of DAM prices for the target time interval \( t \). This two-stage approach aims to enhance the accuracy of predictions by leveraging the error correction capabilities of the FFNN.

\subsection{European Market Structures}

In recent years, the structure and functioning of short-term electricity markets have become increasingly aligned across different regions, with many jurisdictions adopting a similar framework \cite{newbery2016benefits}, largely in response to the rising share of variable Renewable Energy Sources (RES) \cite{silva2022short}. Accordingly, we illustrate the design of a typical European market by examining the configuration of one of the three examples mention in this manuscript, the Greek electricity market.

Over the past decade, the energy sector in Greece has undergone significant changes. These reforms encompass various aspects, such as the liberalization of the wholesale and retail electricity markets, allowing for greater competition. Furthermore, there has been a notable diversification of the electricity generation sources mix, with a significant rise in the proportion of variable RES contributing to the overall final energy consumption, reaching nearly 20\% in 2019 \cite{HenexEnergies} and 25\% in 2023 \cite{Eurostat2023}. Another noteworthy advancement is the transition to the new EU target model market. This includes the establishment of forward, DAM and Intra-day Market (IDM) by the Hellenic Energy Exchange (HEnEx), as well as the introduction of a Balancing Market managed by the Independent Power Transmission Operator (IPTO-ADMIE), in accordance with the EU Target Model. A single distribution system operator, known as the Hellenic Electricity Distribution Network Operator (HEDNO), manages the distribution grid and the non-interconnected islands, whereas IPTO is responsible for the transmission grid. The energy market is overseen by the Regulatory Authority for Energy (RAE).

\begin{figure}[t!]
    \centering
    \includegraphics[width = \columnwidth]{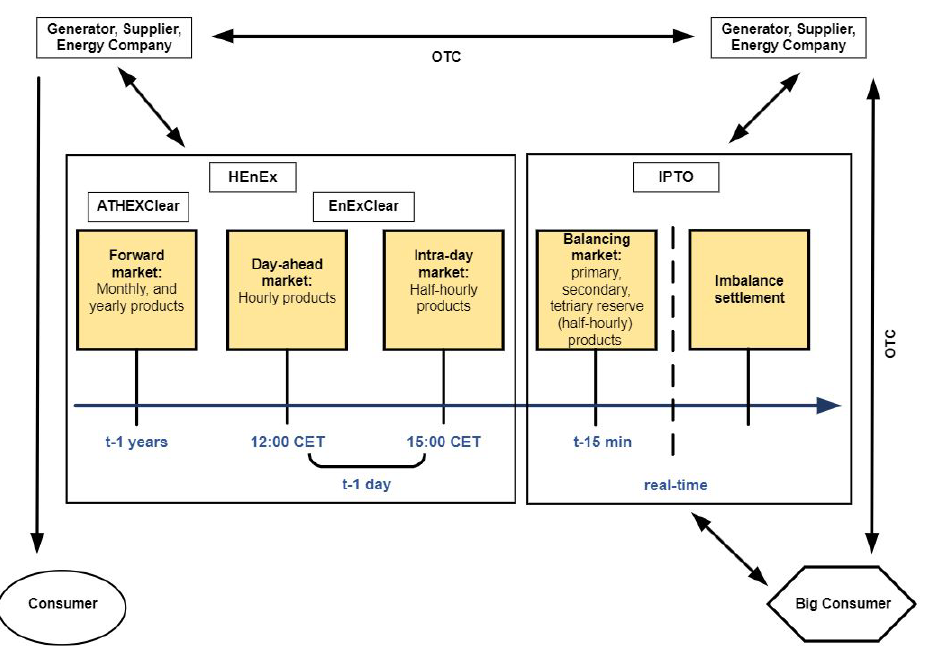}
    \caption{Greek Energy Markets Structure \cite{GreekMarketDesign}}
    \label{fig:GreekMarketDesign}
\end{figure}

The current design of the Greek electricity market is illustrated in Figure \ref{fig:GreekMarketDesign} and is characterized by a centralized framework for trading spot energy, along with separate market mechanisms dedicated to forward and "flexible" capacity trading, in line with the EU Target Model market structure, as follows:

\begin{itemize}
    \item The \textbf{Forward Market (FM)}, operated by HEnEx, enables the trading of both financially and physically settled forward contracts, designed to be monthly or yearly. The transaction clearing for this market is performed by the Athens Exchange Clearing House (ATHEXClear).
    \item \textbf{DAM} tries to match supply and demand on a day-ahead basis. 
    \begin{itemize}
        \item It ensures that all wholesalers and retailers have access to the market and establishes reliable reference system prices (System Marginal Prices).
        \item The product traded is an hourly contract of electricity with physical delivery that specifies the quantity and price, per generator unit.
        \item A Day-Ahead Schedule (DAS) is generated by HEnEx, after solving the EU market price coupling, via an algorithm called EUPHEMIA \cite{enexgroup_pcr}.
    \end{itemize}  
    
    \item \textbf{IDM} market gives participants the opportunity to adjust their positions in close to real-time, particularly in situations involving changes in demand or supply. It also allows for the submission of more precise short-term forecasts for RES.
    \begin{itemize}
        \item Sell and buy hourly orders apply on the day of physical delivery, following the expiration of the DAM deadline
        \item Until 21 September 2021, the Greek IDM performed three local intra-day auctions.
        \item The Greek IDM market is not yet coupled with neighbouring markets in intra-day terms, meaning that the cross-border capacity was not re-optimized after DAM, and thus it was not offered to market participants for trading.
    \end{itemize}
    \item A distinct \textbf{Balancing Market (BM)}, operated by IPTO, comes into play as we approach real-time operations, utilizing a simplified merit-order algorithm to activate Balancing Energy.
\end{itemize}

The \textbf{BM} can be seen as a flexibility trading market, ensuring system safety, as it has not only economic, but mainly physical effects. According to RAE \cite{rae_BM} “the purpose of the Balancing Market is to correct the imbalance between production and demand in real time, maintaining the technical standards of the system and with due account of the results of the previous markets.”

\section{Evaluation}
\label{sec:evaluation}

\subsection{Experimental Setup}

In this study, we evaluate the performance of four distinct ML models: LSTM with FFEC, XGBoost, LightGBM, and CatBoost. Each model is configured with the respected hyperparameters, as summarized in Table \ref{tab:Hyperparams}.

The LSTM with FFEC model leverages recurrent layers to capture temporal dependencies in DAM price series, with FFEC layers refining predictions by correcting residual errors. Hyperparameters such as moderate LSTM units, dropout, and MAE loss were selected to balance learning capacity and overfitting, especially important for short training periods. For the tree-based models (XGBoost, LightGBM, and CatBoost), parameters like learning rate, tree depth, and subsampling were tuned to ensure stable learning and generalization from limited historical data. 

\begin{table}[!t]
\centering
\renewcommand{\arraystretch}{1.5}
\caption{Parameter settings for all models considered}
\label{tab:Hyperparams}
\begin{tabular}{c c c}
\toprule
\makecell[l]{Model} & \makecell[l]{Hyperparameters} & \makecell[l]{Value} \\
\midrule
\multirow{10}{*}{\textbf{LSTM with FFEC}} & \makecell[l]{LSTM, FFEC Optimizers} & \makecell[l]{adam} \\
& \makecell[l]{LSTM Units} & \makecell[l]{200} \\
& \makecell[l]{LSTM Dropout Rate} & \makecell[l]{0.2} \\
& \makecell[l]{LSTM Learning Rate} & \makecell[l]{$1 \times 10^{-4}$} \\
& \makecell[l]{LSTM, FFEC Loss Functions} & \makecell[l]{MAE} \\
& \makecell[l]{LSTM, FFEC Epochs} & \makecell[l]{200} \\
& \makecell[l]{LSTM, FFEC Batch sizes} & \makecell[l]{256} \\
& \makecell[l]{FFEC Layer1,2 Sizes} & \makecell[l]{256, 128} \\
& \makecell[l]{FFEC Layer1,2 Activation Functions} & \makecell[l]{ReLU} \\
\midrule
\multirow{6}{*}{\textbf{XGBoost}}
 & \makecell[l]{Loss Function} & \makecell[l]{MSE} \\ 
& \makecell[l]{Colsample Bytree} & \makecell[l]{0.9} \\ 
& \makecell[l]{Learning Rate} & \makecell[l]{0.01} \\
& \makecell[l]{Tree Maximum Depth} & \makecell[l]{10} \\
& \makecell[l]{Number of Trees} & \makecell[l]{300} \\
& \makecell[l]{Tree Subsample} & \makecell[l]{0.8} \\ 
\midrule
\multirow{6}{*}{\textbf{LightGBM}} & \makecell[l]{Objective} & \makecell[l]{Regression} \\
& \makecell[l]{Colsample Bytree} & \makecell[l]{0.9} \\ 
& \makecell[l]{Learning Rate} & \makecell[l]{0.01} \\
& \makecell[l]{Maximum Depth} & \makecell[l]{10} \\
& \makecell[l]{Number of Trees} & \makecell[l]{300} \\
& \makecell[l]{Subsample} & \makecell[l]{0.8} \\ 
\midrule
\multirow{6}{*}{\textbf{CatBoost}} & \makecell[l]{Objective} & \makecell[l]{RMSE} \\
& \makecell[l]{Learning Rate} & \makecell[l]{0.01} \\
& \makecell[l]{Depth} & \makecell[l]{10} \\
& \makecell[l]{Iterations} & \makecell[l]{500} \\
& \makecell[l]{Subsample} & \makecell[l]{0.8} \\ 
& \makecell[l]{Random Seed} & \makecell[l]{42} \\ 
\bottomrule
\end{tabular}
\end{table}

\subsection{Metrics}

To effectively assess the performance of the models in the DA market, it is essential to measure and compare their respective error variances. To achieve this, a diverse set of widely used metrics has been employed to quantify and compare their accuracy. In this study, the evaluation metrics include the Mean Absolute Error, the Mean Absolute Percentage Error, the Root Mean Squared Error, the Coefficient of Determination ($R^2$), and the Forecast Skill Index (FSI). Each of these metrics captures different aspects of predictive performance, providing complementary insights that contribute to a comprehensive evaluation. This section briefly describes these metrics along with their mathematical formulations.

\begin{table*}[!h]
    \centering
    \caption{Modeling Results}
    \label{tab:results}
    {\renewcommand{\arraystretch}{1.4}
    \resizebox{\textwidth}{!}{%
    \begin{tabular}{p{1.2cm}|p{1.2cm}|ccccc|ccccc|ccccc}
    \toprule
    \multirow{2}{*}{\shortstack{\textbf{Training}\\\textbf{Window}}}
    & \multirow{2}{*}{\shortstack{\textbf{Evaluation}\\\textbf{Metrics}}}
    & \multicolumn{5}{c|}{\textbf{GREECE}} 
    & \multicolumn{5}{c|}{\textbf{BELGIUM}} 
    & \multicolumn{5}{c}{\textbf{IRELAND}} \\ 
    \cline{3-17}
    & & Naive & LSTM & XGB & LGBM & CatBoost & Naive & LSTM & XGB & LGBM & CatBoost & Naive & LSTM & XGB & LGBM & CatBoost \\ 
    \midrule
    \multirow{4}{*}{\textbf{7 days}}  
    & MAE & 26.651 & 36.477 & \textbf{18.561} & 19.167 & 22.123 & 24.171 & 25.838 & \textbf{13.930} & 14.677 & 21.448 & 22.870 & 24.290 & \textbf{12.773} & 13.337 & 17.787 \\ 
    & RMSE & 39.722 & 52.229 & \textbf{27.628} & 27.787 & 33.918 & 34.667 & 34.538 & \textbf{20.792} & 21.424 & 31.008 & 33.808 & 32.486 & \textbf{19.004} & 19.323 & 25.347 \\ 
    & $R^2$ & 0.361 & -0.104 & \textbf{0.691} & 0.687 & 0.534 & 0.437 & 0.441 & \textbf{0.798} & 0.785 & 0.551 & 0.358 & 0.407 & \textbf{0.797} & 0.790 & 0.639 \\ 
    & FSI & 0 & -0.315 & \textbf{0.304} & 0.300 & 0.146 & 0 & 0.004 & \textbf{0.400} & 0.382 & 0.106 & 0 & 0.039 & \textbf{0.438} & 0.428 & 0.250 \\ 
    \midrule
    \multirow{4}{*}{\textbf{14 days}}  
    & MAE & 22.864 & 23.205 & \textbf{15.445} & 15.599 & 18.010 & 23.339 & 19.892 & 12.069 & \textbf{11.300} & 17.361 & 22.821 & 23.647 & 11.883 & \textbf{11.765} & 15.552 \\ 
    & RMSE & 34.133 & 31.532 & 22.980 & \textbf{22.229} & 28.050 & 33.714 & 26.550 & 18.677 & \textbf{17.480} & 26.256 & 32.763 & 36.751 & 19.502 & \textbf{18.837} & 25.004 \\ 
    & $R^2$ & 0.408 & 0.495 & 0.732 & \textbf{0.749} & 0.600 & 0.465 & 0.668 & 0.836 & \textbf{0.855} & 0.674 & 0.428 & 0.280 & 0.797 & \textbf{0.811} & 0.667 \\ 
    & FSI & 0 & 0.076 & 0.327 & \textbf{0.349} & 0.178 & 0 & 0.212 & 0.446 & \textbf{0.482} & 0.221 & 0 & -0.122 & 0.405 & \textbf{0.425} & 0.237  \\ 
    \midrule
    \multirow{4}{*}{\textbf{30 days}}  
    & MAE & 22.546 & 19.104 & 12.700 & \textbf{12.489} & 13.663 & 23.420 & 17.382 & 10.383 & \textbf{9.577} & 12.487 & 22.546 & 17.455 & 9.314 & \textbf{8.538} & 11.682 \\ 
    & RMSE & 33.463 & 29.142 & 19.273 & \textbf{18.653} & 21.251 & 33.348 & 24.074 & 17.756 & \textbf{15.703} & 21.250 & 32.599 & 23.403 & 14.077 & \textbf{12.900} & 18.258 \\ 
    & $R^2$ & 0.292 & 0.463 & 0.765 & \textbf{0.780} & 0.714 & 0.459 & 0.718 & 0.847 & \textbf{0.880} & 0.781 & 0.249 & 0.613 & 0.860 & \textbf{0.882} & 0.765 \\ 
    & FSI & 0 & 0.129 & 0.424 & \textbf{0.443} & 0.365 & 0 & 0.278 & 0.468 & \textbf{0.529} & 0.363 & 0 & 0.282 & 0.568 & \textbf{0.604} & 0.440 \\ 
    \midrule
    \multirow{4}{*}{\textbf{45 days}}  
    & MAE & 25.648 & 17.698 & 13.928 & \textbf{13.020} & 14.311 & 26.274 & 14.998 & 10.037 & \textbf{9.072} & 10.374 & 21.419 & 15.391 & 9.577 & \textbf{8.726} & 11.012 \\ 
    & RMSE & 37.820 & 25.984 & 20.873 & \textbf{19.299} & 22.204 & 36.867 & 20.489 & 16.193 & \textbf{14.441} & 17.699 & 30.742 & 21.375 & 15.544 & \textbf{13.894} & 18.526 \\ 
    & $R^2$ & 0.109 & 0.579 & 0.729 & \textbf{0.768} & 0.693 & 0.255 & 0.771 & 0.857 & \textbf{0.887} & 0.817 & 0.423 & 0.721 & 0.852 & \textbf{0.882} & 0.790 \\ 
    & FSI & 0 & 0.313 & 0.448 & \textbf{0.490} & 0.413 & 0 & 0.444 & 0.561 & \textbf{0.609} & 0.521 & 0 & 0.305 & 0.494 & \textbf{0.548} & 0.397 \\ 
    \midrule
    \multirow{4}{*}{\textbf{60 days}}  
    & MAE & 21.121 & 18.839 & 18.303 & \textbf{12.030} & 12.827 & 23.702 & 13.324 & 8.830 & \textbf{7.921} & 10.075 & 22.023 & 12.982 & 8.253 & \textbf{7.123}\textsuperscript{\ding{72}} & 9.273 \\ 
    & RMSE & 30.255 & 41.739 & \textbf{12.399}\textsuperscript{\ding{72}} & 17.492 & 19.252 & 33.885 & 18.631 & 14.153 & \textbf{12.457}\textsuperscript{\ding{72}} & 17.098 & 31.569 & 18.126 & 12.780 & \textbf{10.917}\textsuperscript{\ding{72}} & 14.627 \\ 
    & $R^2$ & 0.338 & -0.260 & 0.758 & \textbf{0.779} & 0.732 & 0.347 & 0.803 & 0.886 & \textbf{0.912}\textsuperscript{\ding{72}} & 0.834 & 0.387 & 0.798 & 0.900 & \textbf{0.927}\textsuperscript{\ding{72}} & 0.868 \\ 
    & FSI & 0 & -0.380 & 0.395 & \textbf{0.422} & 0.364 & 0 & 0.450 & 0.582 & \textbf{0.632} & 0.495 & 0 & 0.426 & 0.595 & \textbf{0.654}\textsuperscript{\ding{72}} & 0.537  \\ 
    \midrule
    \multirow{4}{*}{\textbf{90 days}}  
    & MAE & 22.454 & 14.636 & 12.949 & \textbf{11.899}\textsuperscript{\ding{72}} & 12.979 & 23.739 & 12.269 & 8.385 & \textbf{7.365}\textsuperscript{\ding{72}} & 8.806 & 23.451 & 12.989 & 8.423 & \textbf{7.485} & 8.719 \\ 
    & RMSE & 34.519 & 22.199 & 20.658 & \textbf{17.891} & 21.165 & 35.634 & 18.323 & 14.368 & \textbf{12.805} & 15.195 & 33.756 & 18.374 & 13.300 & \textbf{11.701} & 14.111 \\ 
    & $R^2$ & 0.320 & 0.719 & 0.756 & \textbf{0.817}\textsuperscript{\ding{72}} & 0.744 & 0.260 & 0.804 & 0.880 & \textbf{0.904} & 0.865 & 0.144 & 0.746 & 0.867 & \textbf{0.897} & 0.850 \\ 
    & FSI & 0 & 0.357 & 0.402 & \textbf{0.482}\textsuperscript{\ding{72}} & 0.387 & 0 & 0.486 & 0.597 & \textbf{0.641}\textsuperscript{\ding{72}} & 0.574 & 0 & 0.456 & 0.606 & \textbf{0.653} & 0.582 \\ 
    \bottomrule
\end{tabular}%
}}
\end{table*}

The MAE is one of the most commonly used metrics, measuring the average absolute difference between predicted and actual values. It is scale-dependent, expressed as a single numerical value, and treats all errors equally, providing a general indication of a model’s performance. Its mathematical formulation is:  

\begin{equation}
    MAE = \frac{1}{T} \sum_{t=1}^{T} \left| p_t - a_t \right|
\end{equation}

The MAPE, on the other hand, converts the mean absolute error into a percentage, offering a relative measure that is independent of scale. However, it may not always be fully comparable, as the denominator can vary significantly. Its equation is given by:

\begin{equation}
    MAPE = \frac{1}{T} \sum_{t=1}^{T} \frac{\left| p_t - a_t \right|}{a_t} \times 100\%
\end{equation}

The RMSE emphasizes larger errors by squaring the differences before averaging them, making it particularly useful for assessing a model’s robustness. A lower RMSE value indicates higher forecasting accuracy. The formula is:

\begin{equation}
    RMSE = \sqrt{\frac{1}{T} \sum_{t=1}^{T} (p_t - a_t)^2}
\end{equation}

The $R^2$ score evaluates a model’s ability to explain the variance in actual energy production. It measures the proportion of variance captured by the forecasting model, with values ranging from 0 (no explanatory power) to 1 (perfect prediction). The formula is:

\begin{equation}
    R^2 = 1 - \frac{\sum_{t=1}^{T} (p_t - a_t)^2}{\sum_{t=1}^{T} (\bar{a} - a_t)^2}
\end{equation}

While these traditional error metrics—RMSE, MAE, MAPE, and $R^2$—are widely used to evaluate forecasting models, they may not be sufficient when comparing models across different datasets, locations, or forecasting horizons \cite{ilias2023unsupervised}. In this context, the FSI provides an additional layer of evaluation by benchmarking model accuracy against a naive persistence model. This approach ensures a more objective assessment of predictive capabilities. Mathematically, FSI is defined as:

\begin{equation}
    FSI = 1 - \frac{RMSE_{\text{model}}}{RMSE_{\text{persistence}}}
\end{equation}

where $RMSE_{\text{persistence}}$ corresponds to the relevant error of a naive persistence model, which assumes that future values remain unchanged from the most recent observed data. The choice of the reference model in forecast skill evaluation plays a crucial role in performance assessment, and in this study, the naive model is used as a baseline. Incorporating FSI alongside traditional error metrics, provides a more well-rounded evaluation of forecasting accuracy in the DAM.

\begin{figure*}[!t]
    \centering
    \begin{minipage}{0.3\textwidth}
        \centering
        \includegraphics[width=\textwidth]{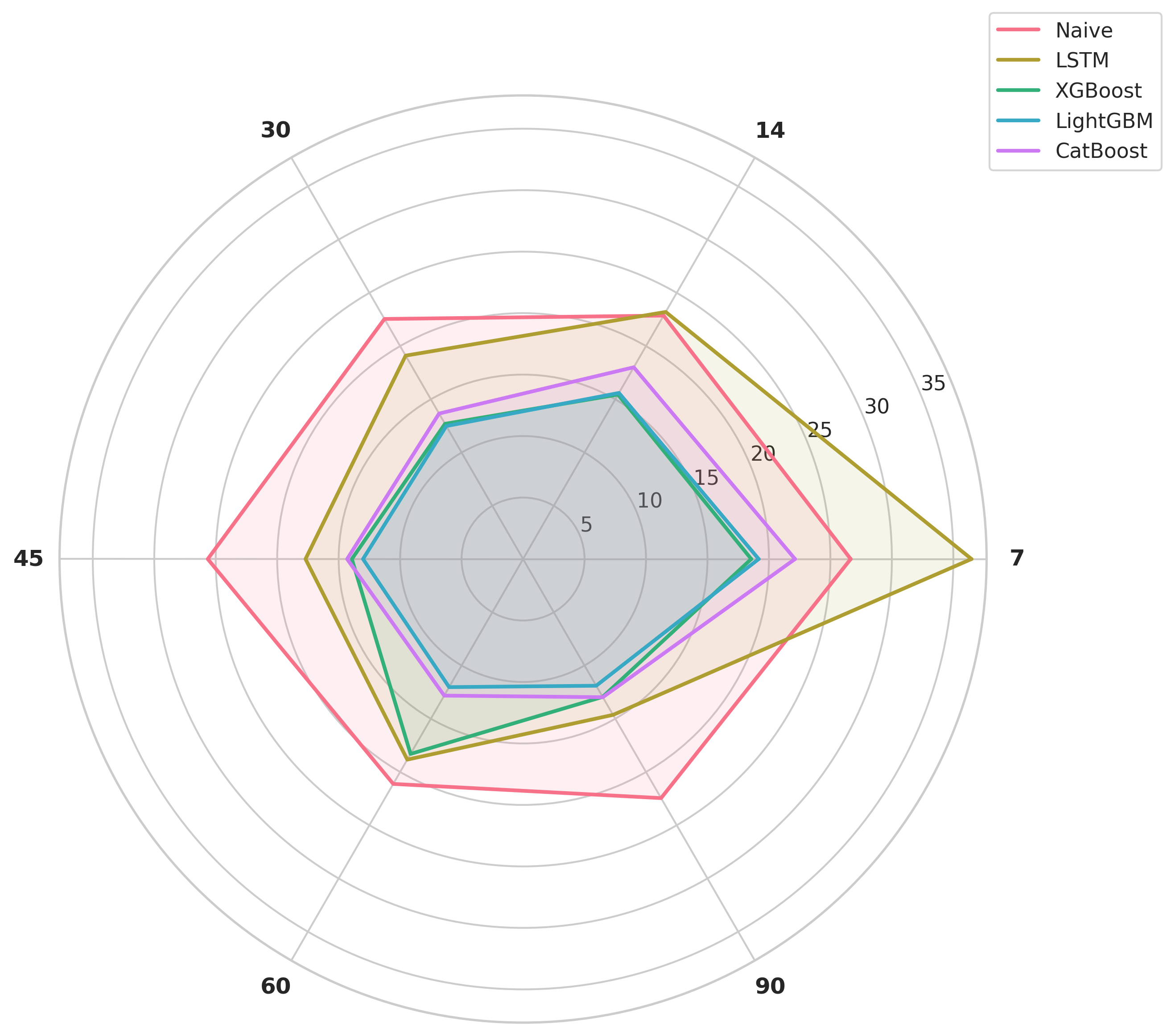}
        
        \vspace{1mm}
        \small{(a) Greece}
    \end{minipage}
    \hfill
    \begin{minipage}{0.3\textwidth}
        \centering
        \includegraphics[width=\textwidth]{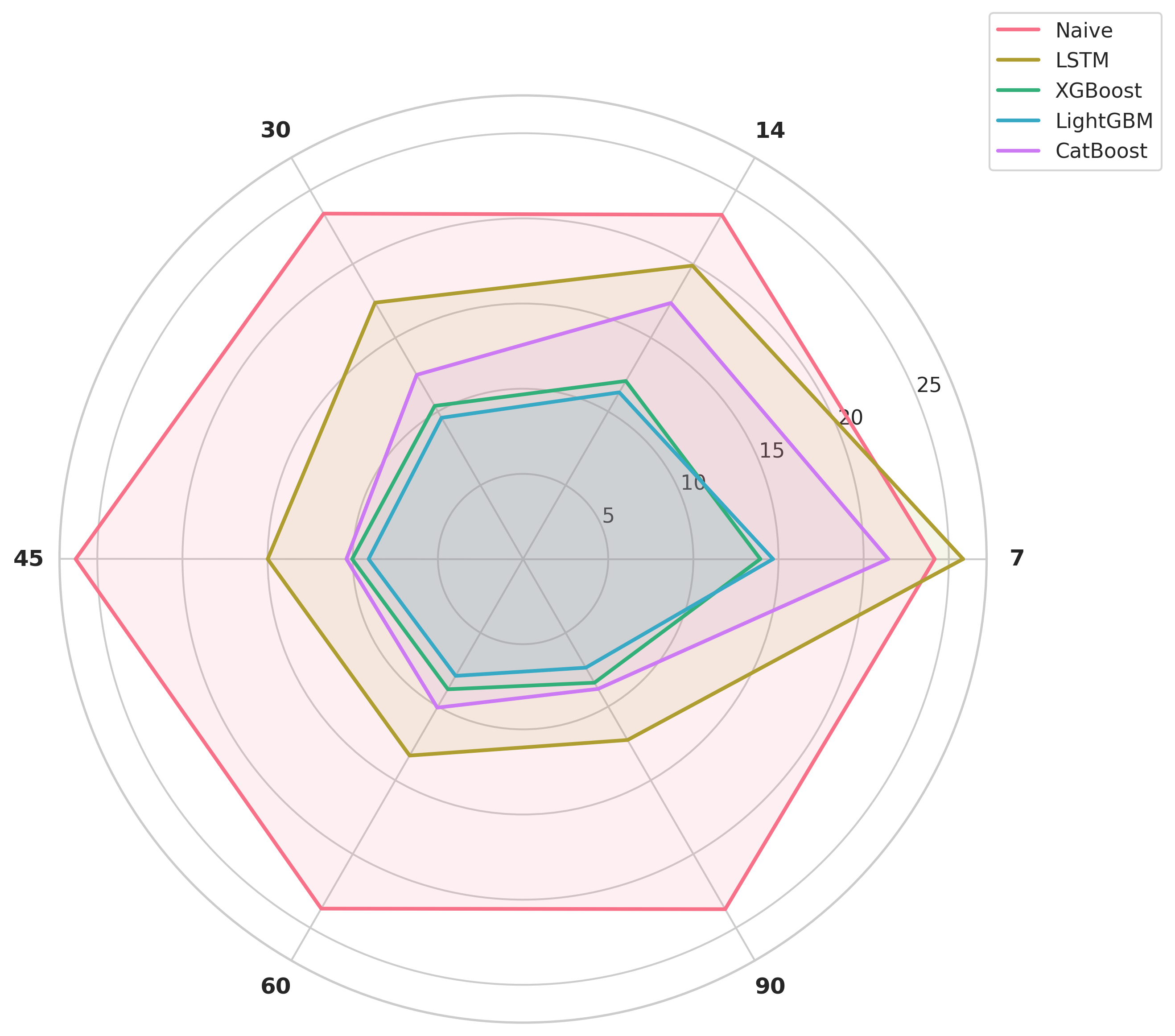}
        
        \vspace{1mm}
        \small{(b) Belgium}
    \end{minipage}
    \hfill
    \begin{minipage}{0.3\textwidth}
        \centering
        \includegraphics[width=\textwidth]{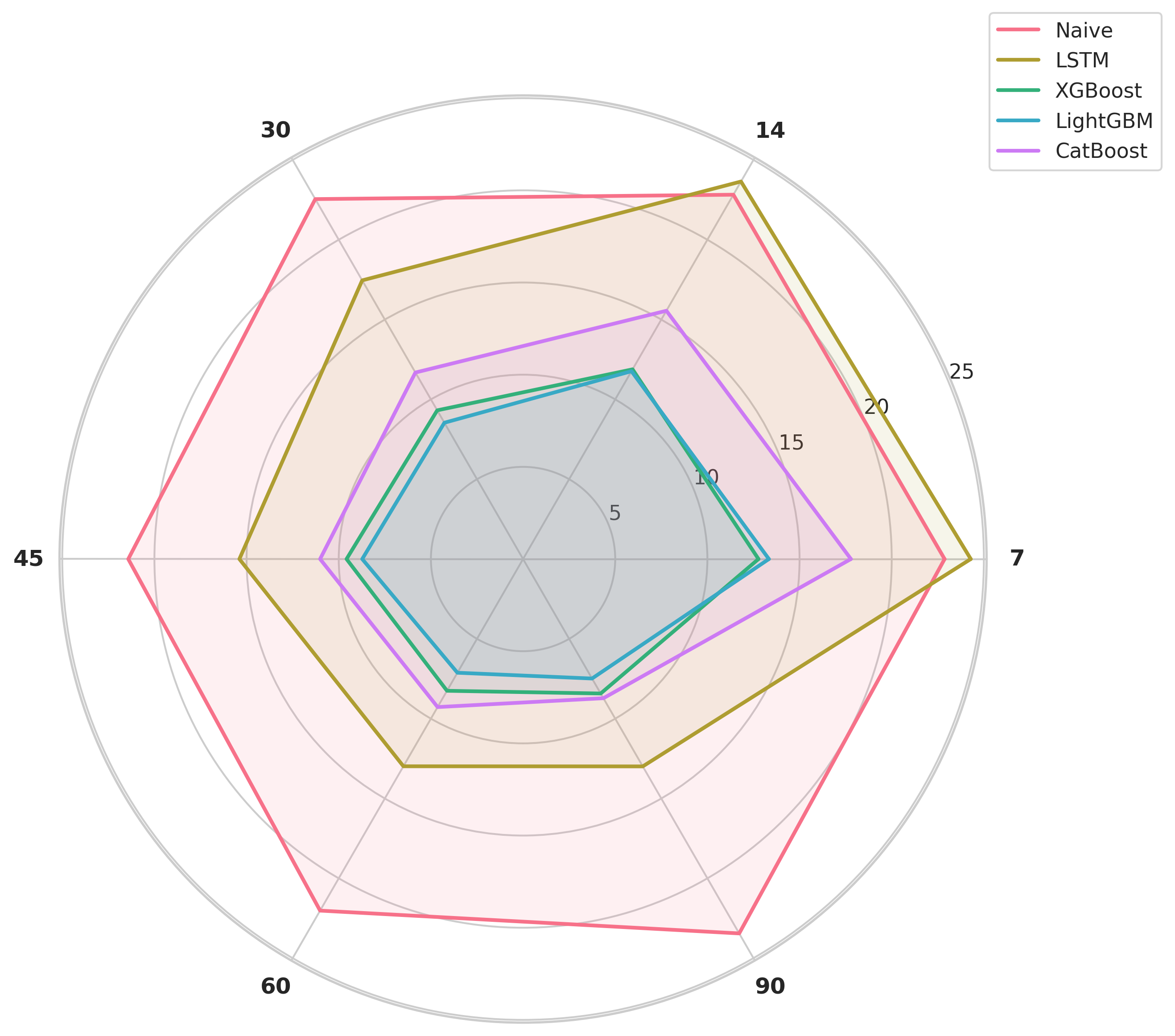}
        
        \vspace{1mm}
        \small{(c) Ireland}
    \end{minipage}

    \caption{Radar plots comparing model performance across different training windows (7, 14, 30, 45, 60, and 90 days) for Greece, Belgium, and Ireland. Each axis represents a training window, and lower values indicate better performance.}
    \label{fig:radar_plots}
\end{figure*}

\subsection{Cross-model Comparative Analysis}

In this study, five models were evaluated for DAM price forecasting: Naive, LSTM, XGBoost, LightGBM, and CatBoost. Table \ref{tab:results} presents the experimental results for each of the three countries across different training windows (7, 14, 30, 45, 60, and 90 days). In bold are the best metrics for each training window and assigned with \textsuperscript{\ding{72}} are the optimal ones for all training windows for each country. As observed, LightGBM consistently outperforms the other models in nearly every metric and training window. Notably, it achieves lower MAE and RMSE values while maintaining high $R^2$ and FSI scores across all three countries. Moreover, LightGBM exhibits superior computational efficiency, with significantly lower training times compared to LSTM and the other boosting models. 

The impact of the training window size is evident, as increasing the training history generally leads to better model performance. This is reflected in the decreasing MAE and RMSE values and the increasing $R^2$ scores. Shorter training windows (7 and 14 days) tend to yield lower accuracy, whereas longer windows, particularly 90 days, result in more reliable forecasts. However, in some cases, the 60-day training window achieves better results than the 90-day window, especially for RMSE (e.g., Greece and Ireland for LightGBM). This suggests that while longer training windows generally improve forecasting performance, an optimal window length may exist, beyond which additional historical data introduces noise or reduces adaptability to recent market trends.

Among the other models, XGBoost and CatBoost perform well but consistently fall short of LightGBM. LSTM, on the other hand, struggles with shorter training windows, sometimes producing negative $R^2$ values (e.g., Greece, 7 days, $R^2 = -0.104$), indicating weak predictive performance. This suggests that while LSTMs are well-suited for sequential modeling, they may require extensive hyperparameter tuning and longer training histories to achieve competitive results.

The FSI further supports the superiority of LightGBM, as it achieves the highest FSI scores across most cases. The FSI values increase with the training window length, reinforcing the observation that more historical data leads to better forecasting accuracy. While XGBoost and CatBoost also show high FSI values, LightGBM remains the best-performing model overall. These findings confirm that LightGBM is the most effective and efficient approach for DAM price forecasting, highlighting the importance of training window selection in improving predictive performance. 


The radar plots in Figure \ref{fig:radar_plots} visually complement the results presented in Table \ref{tab:results}, offering a clear comparison of model performance across different training windows for each country. The plots illustrate how the prediction error varies with different training periods, with lower values indicating better performance. LightGBM consistently achieves the lowest error in most cases, reinforcing its dominance in the tabular results. However, XGBoost and CatBoost also demonstrate competitive performance, particularly for longer training windows. Notably, in some instances, the 60-day training window yields better results than the 90-day window, suggesting that an optimal balance between training data size and model generalization exists. The plots further highlight the relatively weaker performance of LSTM and the Naive model, particularly for shorter training windows, emphasizing the superiority of gradient boosting methods for DAM price forecasting.

\subsection{Seasonal \& Market Price Spike Assessment}

\begin{figure*}[!t]
    \centering
    \begin{minipage}{0.3\textwidth}
        \centering
        \includegraphics[width=\textwidth]{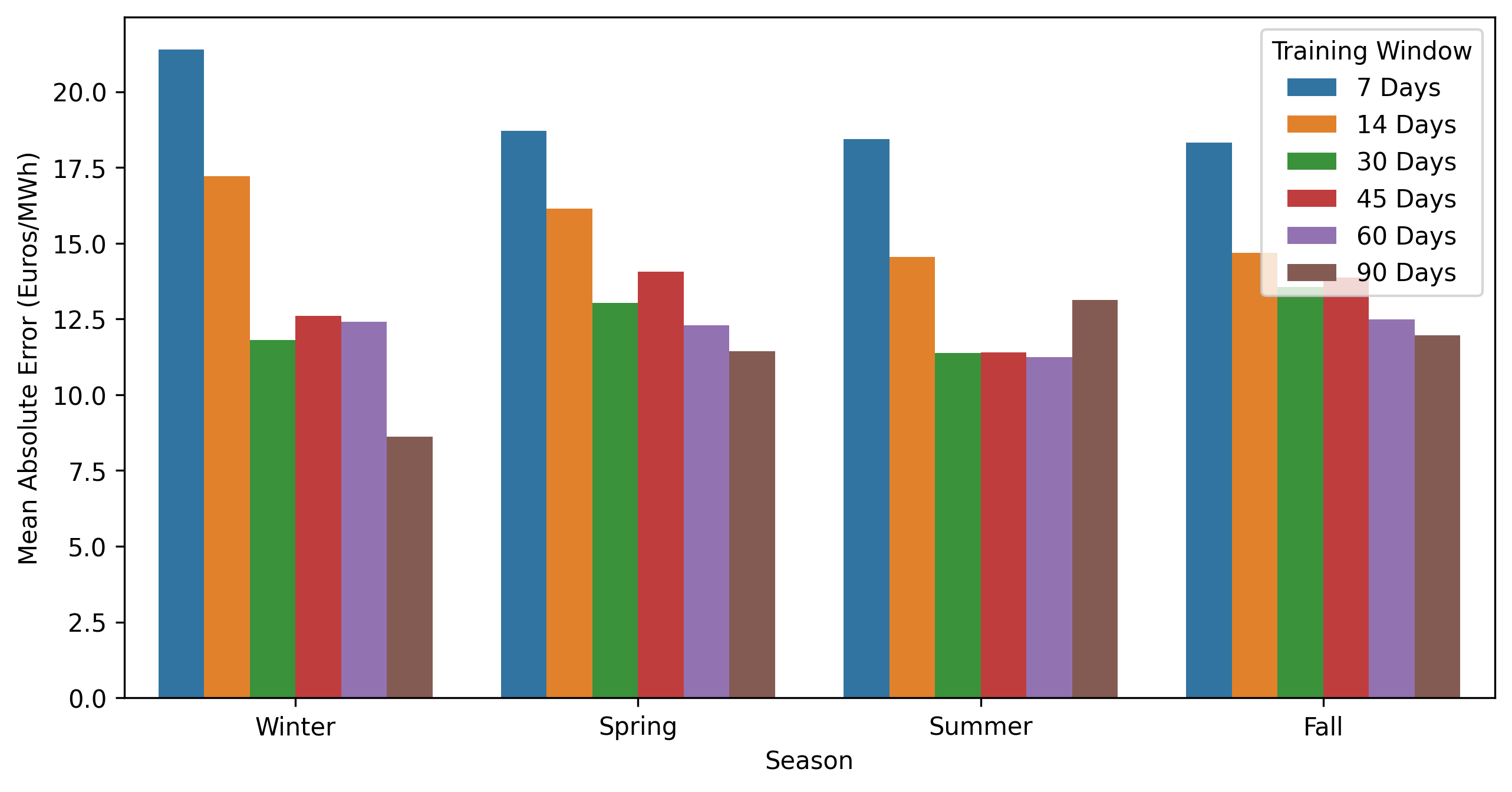}
        
        \vspace{1mm}
        \small{(a) Greece}
    \end{minipage}
    \hfill
    \begin{minipage}{0.3\textwidth}
        \centering
        \includegraphics[width=\textwidth]{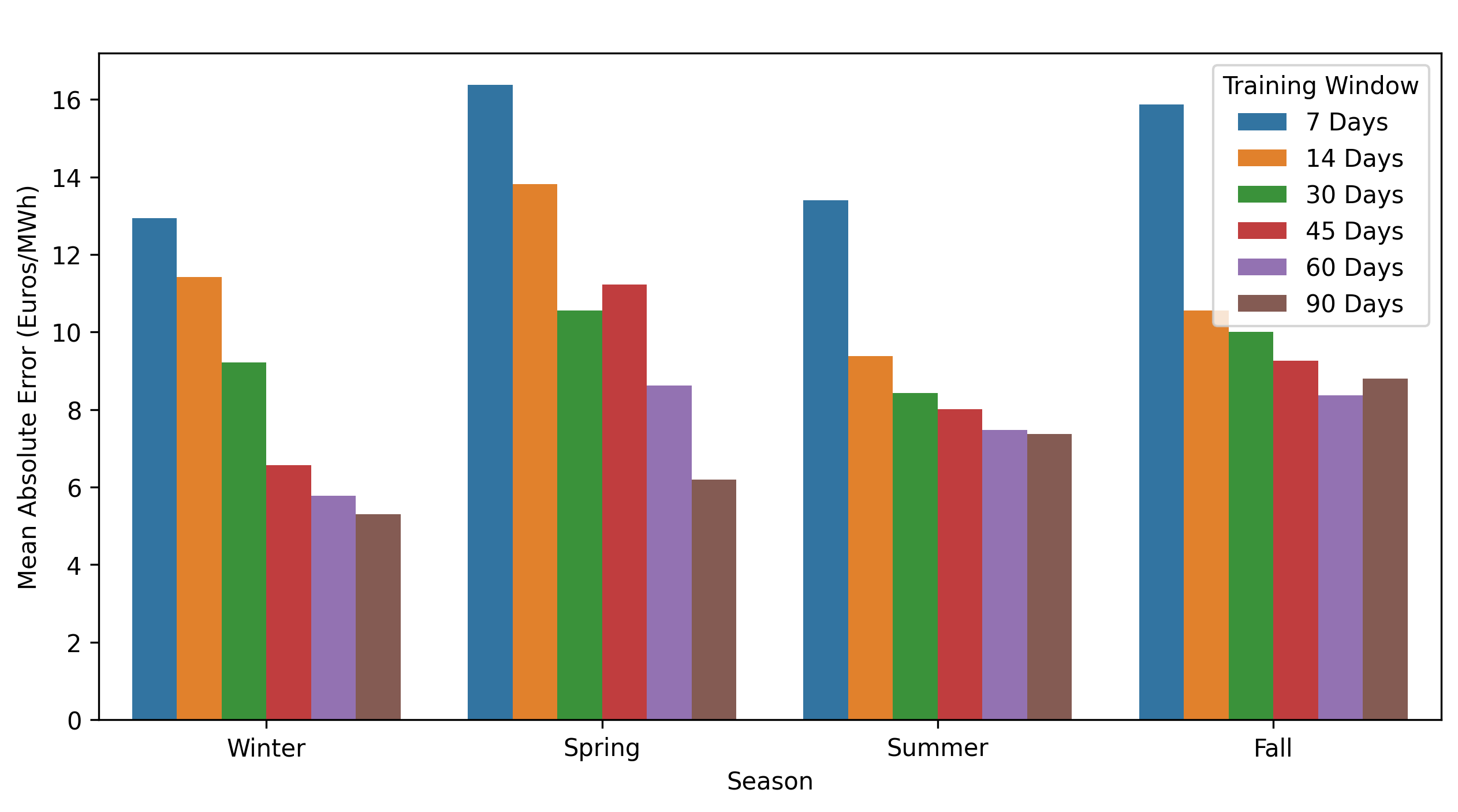}
        
        \vspace{1mm}
        \small{(b) Belgium}
    \end{minipage}
    \hfill
    \begin{minipage}{0.3\textwidth}
        \centering
        \includegraphics[width=\textwidth]{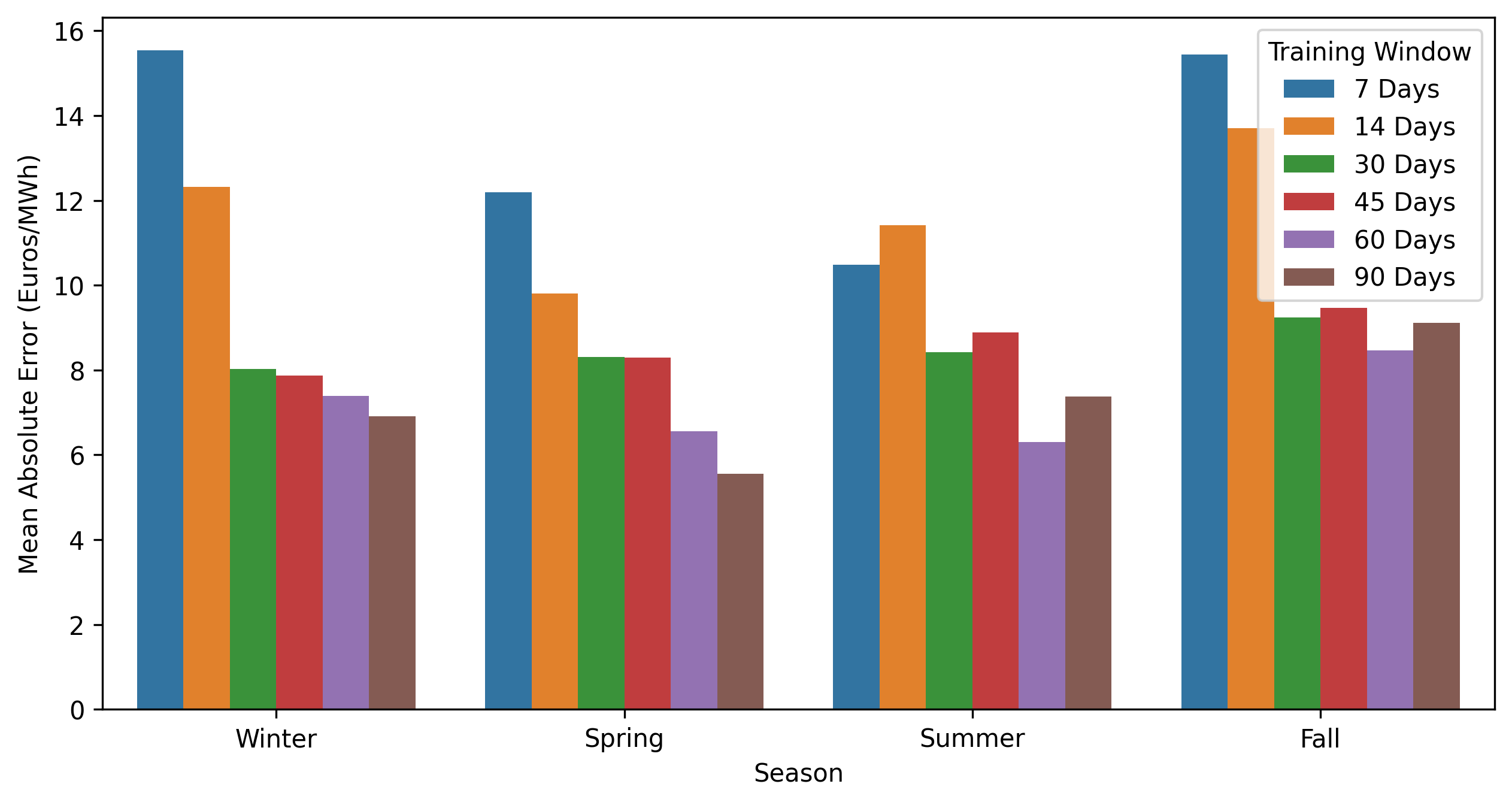}
        
        \vspace{1mm}
        \small{(c) Ireland}
    \end{minipage}

    \caption{Seasonal MAE for different training windows across Greece, Belgium, and Ireland}
    \label{fig:seasonal_plots}
\end{figure*}

\begin{figure*}[!t]
    \centering
    \begin{minipage}{0.3\textwidth}
        \centering
        \includegraphics[width=\textwidth]{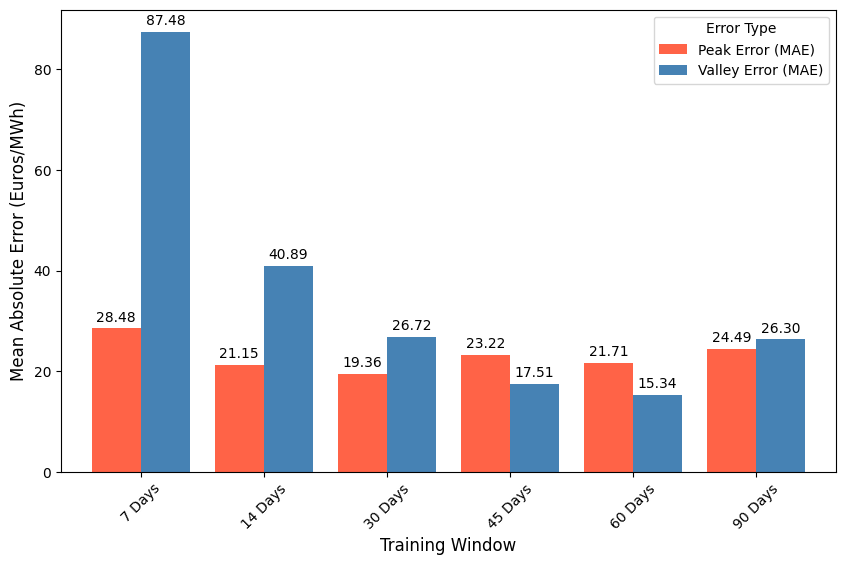}
        
        \vspace{1mm}
        \small{(a) Greece}
    \end{minipage}
    \hfill
    \begin{minipage}{0.3\textwidth}
        \centering
        \includegraphics[width=\textwidth]{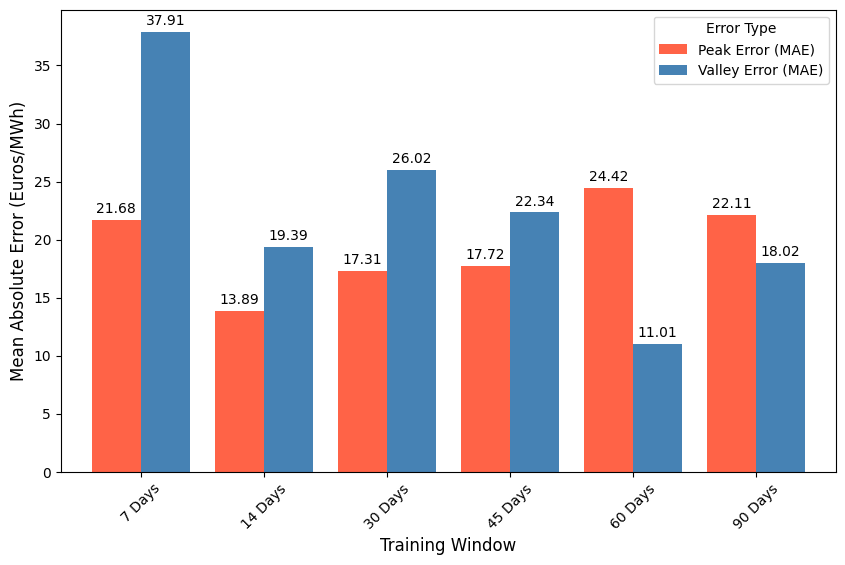}
        
        \vspace{1mm}
        \small{(b) Belgium}
    \end{minipage}
    \hfill
    \begin{minipage}{0.3\textwidth}
        \centering
        \includegraphics[width=\textwidth]{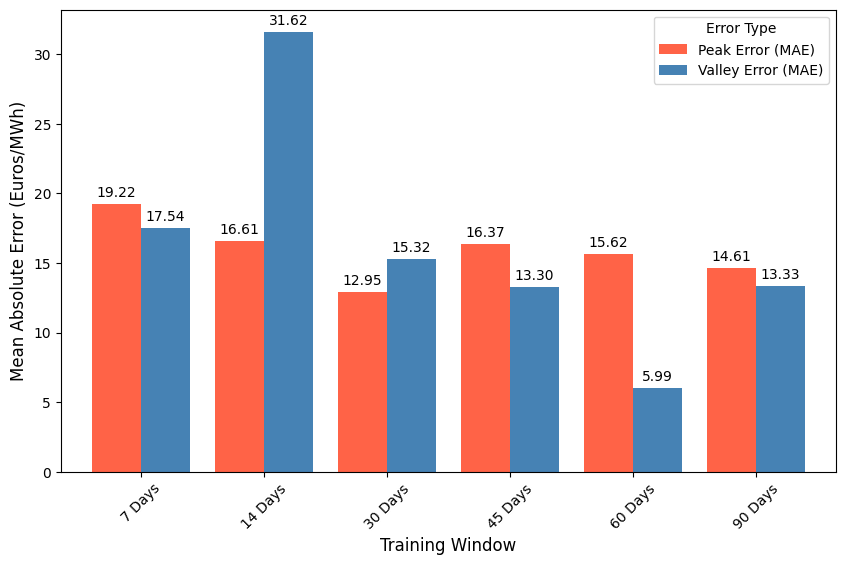}
        
        \vspace{1mm}
        \small{(c) Ireland}
    \end{minipage}

    \caption{Peak analysis for different countries: (a) Greece, (b) Belgium, and (c) Ireland.}
    \label{fig:peak_plots}
\end{figure*}

Electricity market prices are influenced by a variety of factors, including seasonal demand variations, supply constraints, and external economic conditions. In addition to evaluating model performance across different training windows, it is crucial to assess how well these models capture seasonal fluctuations and sudden price spikes. Such extreme price deviations can significantly impact market participants, necessitating models that not only provide accurate forecasts under normal conditions but also respond effectively to rapid market changes. This section analyzes the ability of the evaluated models to adapt to seasonal trends and predict extreme price movements, offering insights into their robustness and reliability in real-world scenarios.

In Figure \ref{fig:seasonal_plots}, the seasonal MAE plots provide insights into the influence of seasonality on forecasting accuracy across different training window lengths. In general, shorter training windows (e.g., 7 or 14 days) exhibit higher errors again, particularly during winter and spring, likely due to increased market volatility. Longer training windows (90 days) consistently demonstrate lower MAE across seasons for all countries, highlighting their ability to capture long-term trends.

In general, summer and fall are the seasons with the highest fluctuations in energy markets, driven by factors such as increased demand during hot temperatures, changes in energy production due to weather conditions and shifts in supply from RES. These fluctuations are particularly pronounced during peak consumption periods, making accurate forecasting critical for market participants. Interestingly, the 45-day and 60-day training windows tend to outperform the 90-day window in these seasons. The 90-day window incorporates a broader range of data, including less relevant or outdated trends that might not accurately reflect the current market dynamics. As the energy market responds more strongly to short-term fluctuations in demand and supply, the shorter 45-day and 60-day windows are better able to capture these seasonal variations and provide more responsive predictions. These results emphasize the importance of seasonality in selecting appropriate forecasting strategies.

Delving deeper into our analysis, it becomes evident that the choice of training window significantly impacts the accuracy of peak and valley predictions across different countries. As illustrated in Figure \ref{fig:peak_plots}, both Greece and Belgium exhibit notably higher errors with the 7-day window, especially for valley predictions, reflecting the insufficiency of shorter training periods in capturing complex market behaviors. In contrast, the 45-day and 60-day training windows consistently yield lower MAE values for both peak and valley forecasts, indicating a more balanced responsiveness to recent trends without the dilution effect seen in the 90-day window. This pattern is particularly clear in countries like Greece, where the 30 and 60-day window achieves the lowest peak error of all configurations. These findings reinforce the hypothesis that medium-range training windows—specifically 45 and 60 days—strike an optimal balance by incorporating relevant, recent market data while excluding outdated information, thereby enhancing the forecasting model’s adaptability to seasonal volatility and sudden market shifts typical of summer and fall periods.

\section{Conclusion and Outlook}
\label{sec:conclusions}

Concluding, accurate forecasting of DAM electricity prices is critical for the efficient operation of power systems and the financial planning of market participants. As the growing share of RES and increased cross-border trading amplify market volatility, numerous studies have sought to address these challenges. This paper contributes to this effort by evaluating ML approaches to enhance price forecasting accuracy across three European markets, namely Greece, Belgium and Ireland, using shallow training window sizes and a careful data pre-processing.

More specifically, a comprehensive evaluation of an LSTM with FFEC, XGBoost, LightGBM, and CatBoost models over varying restricted training windows (7, 14, 30, 45, 60 and 90 days), we identified clear patterns in model performance and data sensitivity. The findings highlight that LightGBM consistently outperforms other models in terms of both accuracy and computational efficiency, especially when using medium-range training windows (45 to 60 days). These windows strike an optimal balance between capturing relevant historical trends and avoiding the inclusion of outdated or noisy data, as reflected in both overall error metrics and the models' ability to predict seasonal fluctuations and price spikes. Furthermore, the importance of careful data pre-processing is underscored. The use of forecasted features from ENTSO-E, rather than actual historical values, provided models with realistic inputs aligned with operational forecasting conditions. Historical forecast data for electricity demand, RES generation, total generation, and transmission net flows were also collected to serve as additional features, aiming at improving DAM price prediction accuracy. Additionally, time-step shifting proved critical for preparing datasets, particularly for models sensitive to input scale and temporal sequences.

Looking ahead, as the EU moves toward a more interconnected and decarbonized energy system, forecasting approaches will need to support greater spatial granularity, shorter decision-making cycles, and alignment with regulatory frameworks promoting transparency and flexibility. Investment in scalable, adaptive ML systems will be crucial to meeting these emerging needs, enabling stakeholders to navigate an increasingly dynamic electricity landscape with confidence whilst increasing both profitability and sustainability.


\bibliographystyle{unsrt}
\bibliography{ref}

\begin{IEEEbiography}
[{\includegraphics[width=1in,height=1.25in,clip,keepaspectratio]{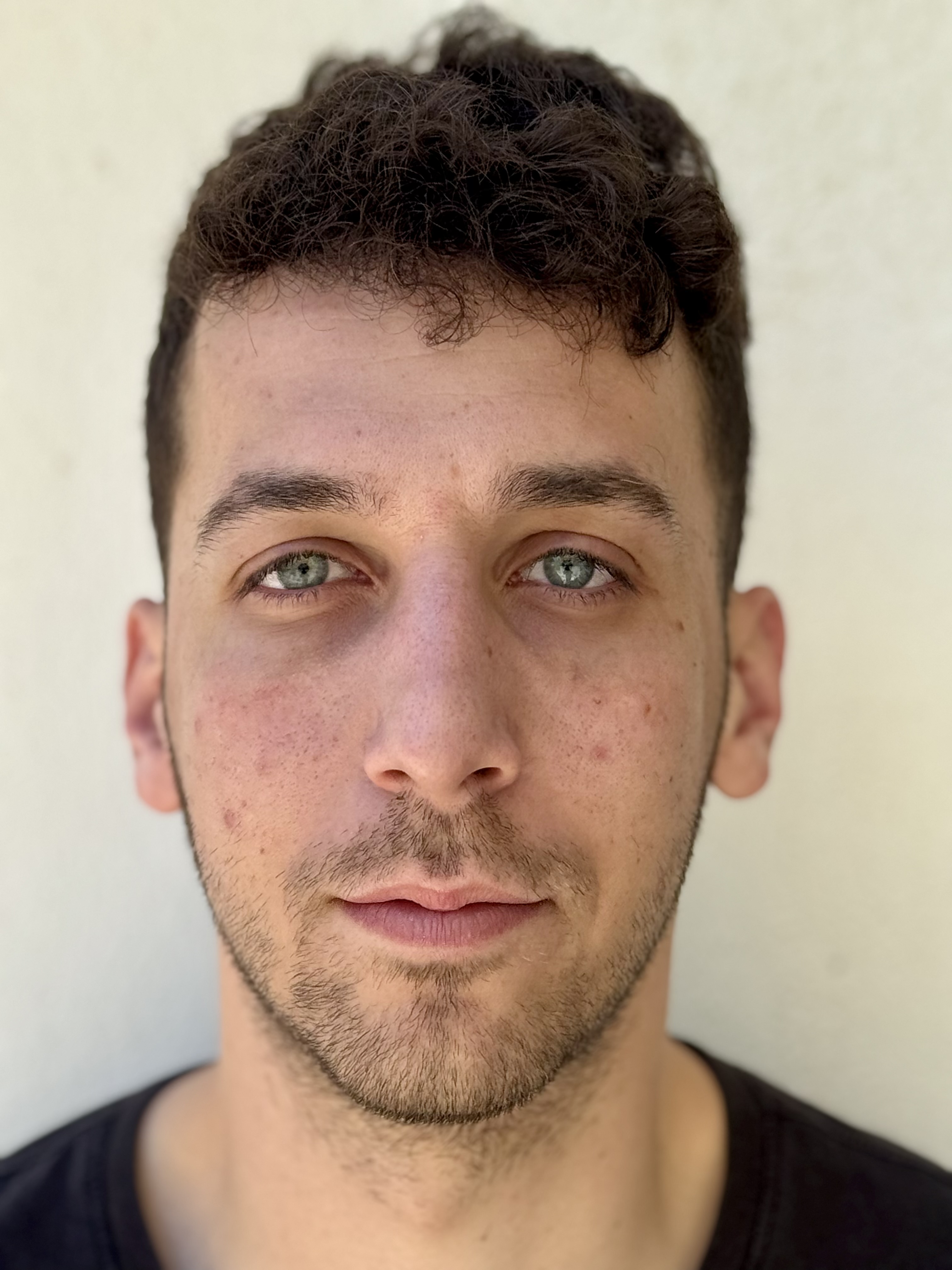}}]{Vasilis Michalakopoulos} is currently a Ph.D. candidate at the National Technical University of Athens (NTUA), researching artificial intelligence applications in energy management, with a focus on demand response optimization and interaction with energy markets. In parallel, he works as a Research Associate on multiple EU-funded projects, contributing to the design and development of data-driven systems for energy and sustainability challenges. He holds a Diploma (integrated BSc and MEng) in Electrical and Computer Engineering from NTUA (2021) and brings a strong background in DevOps engineering.
\end{IEEEbiography}

\begin{IEEEbiography}
[{\includegraphics[width=1in,height=1.25in,clip,keepaspectratio]{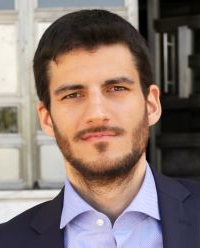}}]{Christoforos Menos-Aikateriniadis} received the Dipl.-Ing degree in electrical and computer engineering from the Aristotle University of Thessaloniki, Thessaloniki, Greece, in 2017, and the M.Sc. degree in sustainable energy systems from Universitat Politècnica de Catalunya (UPC) Barcelona, Spain and KTH Royal Institute of Technology, Stockholm, Sweden, in 2019. During 2019-2021 he worked as an energy markets consultant at AFRY Management Consulting, Oxford, UK. Christoforos is currently pursuing the Ph.D. degree in electrical and computer engineering at the National Technical University of Athens, Greece, working data-driven methods for the Integration of flexible
loads in demand-side management schemes. His research is funded by the Marie Skłodowska Curie Actions (MSCA) project GECKO. His interests include demand-side management, energy markets, optimal scheduling of distributed energy resources, and AI-aided smart grid solutions. 
\end{IEEEbiography}

\begin{IEEEbiography}
[{\includegraphics[width=1in,height=1.25in,clip,keepaspectratio]{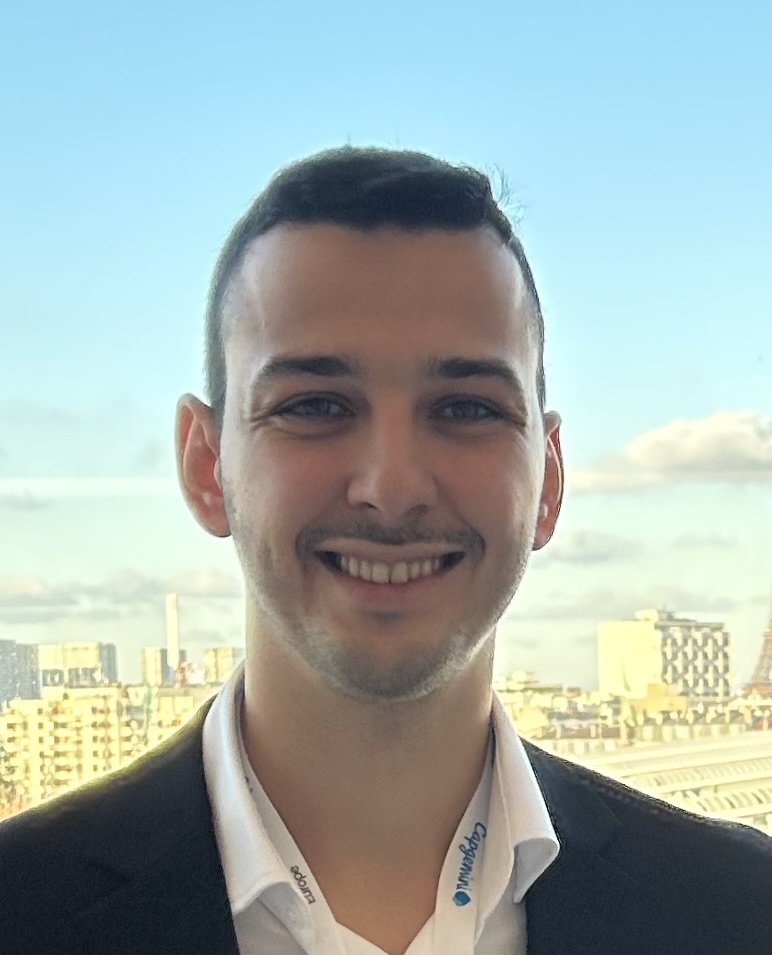}}]{Elissaios Sarmas} is a Senior Research Associate with the National
Technical University of Athens (NTUA), specializing in the integration of
artificial intelligence within energy systems. His research encompasses the
development of machine learning models for energy management, focusing
on renewable energy forecasting, demand response optimization, and
enhancing flexibility in microgrids. He has contributed to several European
research projects, delivering innovative solutions for energy efficiency
and smart grid applications. His scholarly work includes publications on
transfer learning strategies for solar power forecasting and the application of
meta-learning in photovoltaic power prediction, reflecting his commitment
to advancing intelligent energy management systems.
\end{IEEEbiography}

\begin{IEEEbiography}[{\includegraphics[width=1in,height=1.25in,clip,keepaspectratio]{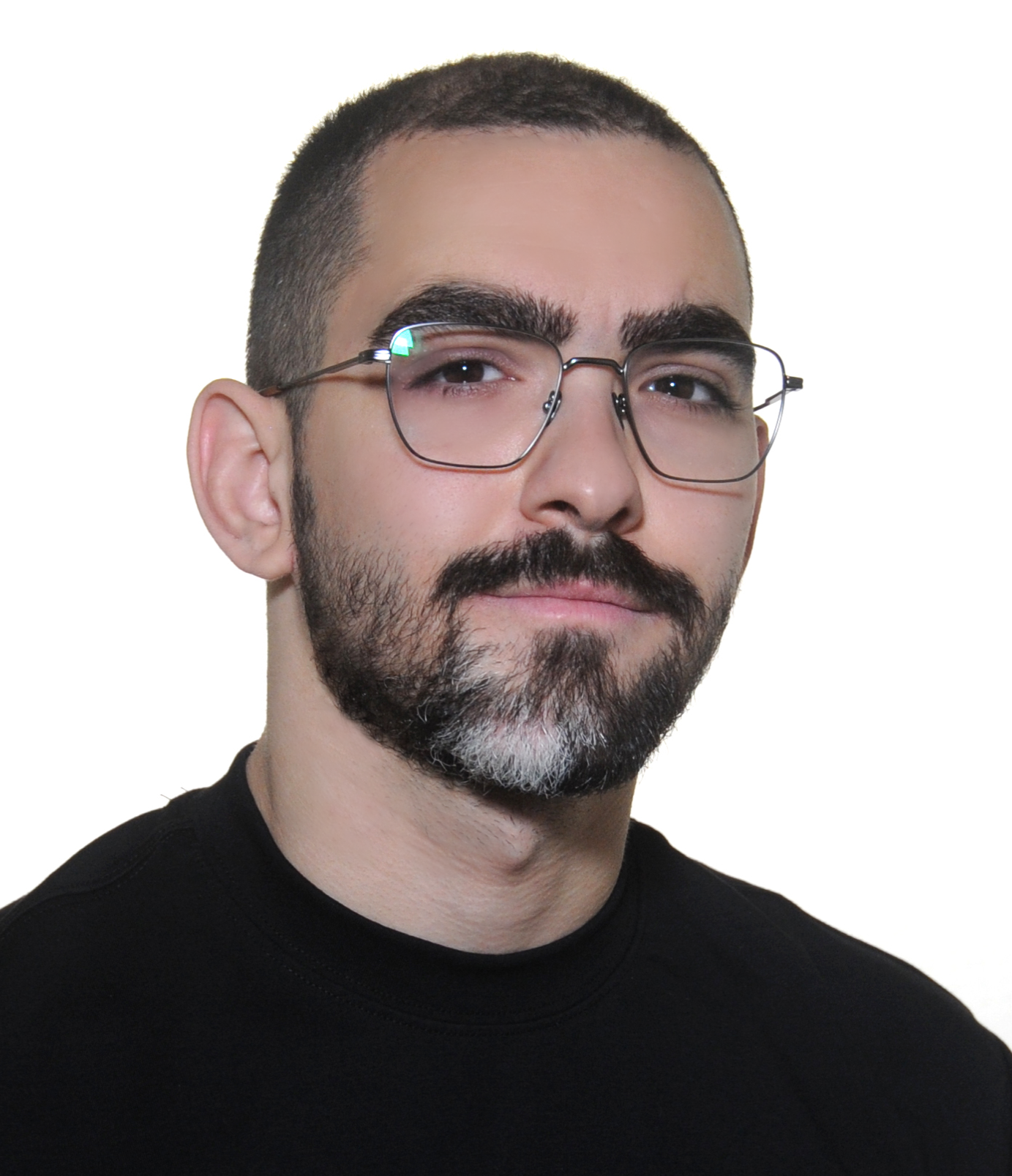}}]{Antonis Zakynthinos} received his Diploma (BS and MEng) in Electrical and Computer Engineering from the National Technical University of Athens (NTUA) in 2025. Since 2024, he has been working as a Machine Learning Engineer on EU-funded research projects, developing AI-powered solutions for challenges related to the energy sector. Concurrently, he has been engaged as a Research Associate at NTUA, with research interests centered on artificial intelligence applications in the energy sector, particularly in the areas of forecasting, operational optimization, and smart grid integration.
\end{IEEEbiography}

\begin{IEEEbiography}[{\includegraphics[width=1.1in,height=1.25in,clip,keepaspectratio]{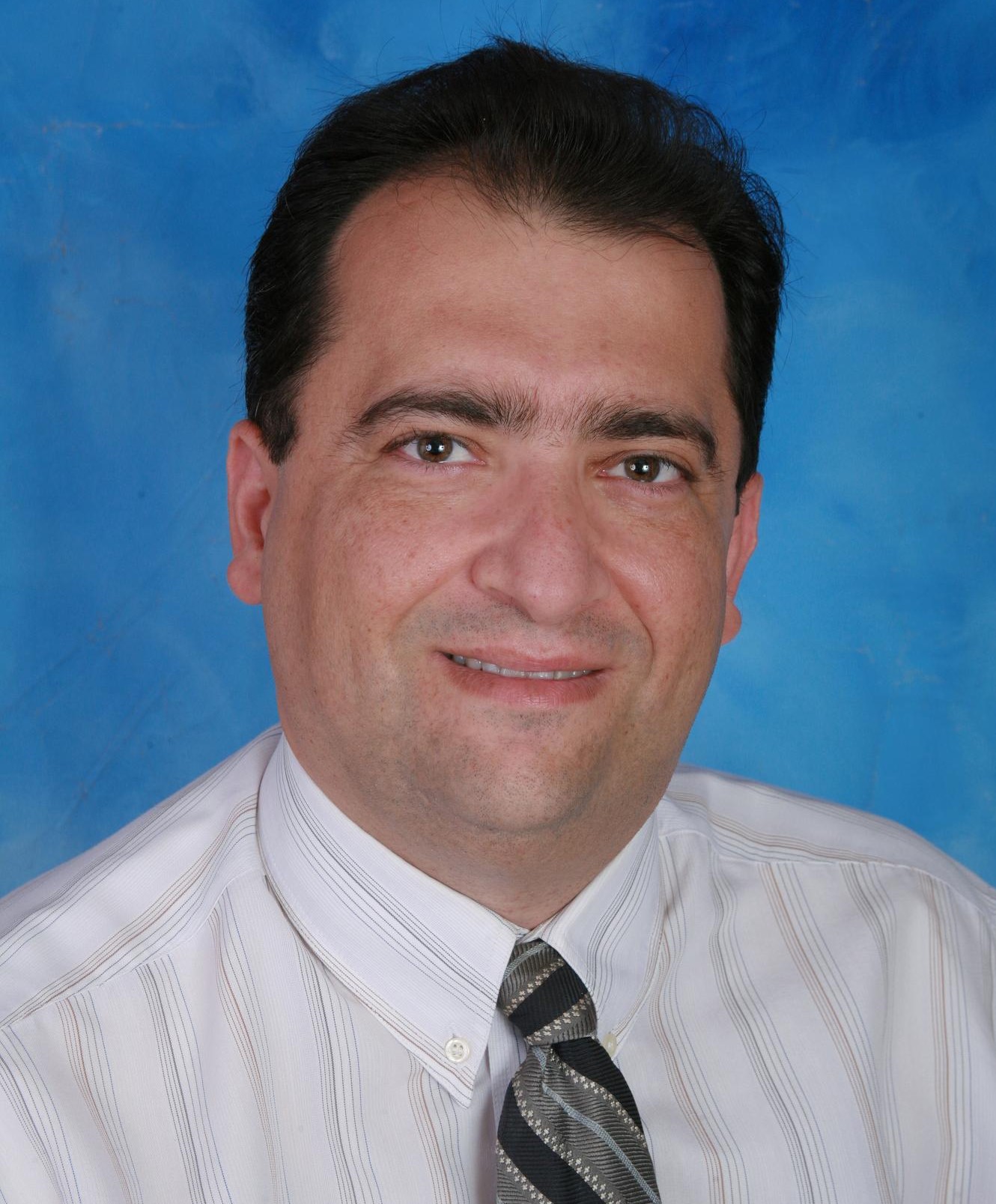}}]{PAVLOS S. GEORGILAKIS}(Senior Member, IEEE) received the Diploma and Ph.D. degrees in electrical and computer engineering from the National Technical University of Athens (NTUA), Athens, Greece, in 1990 and 2000, respectively. In September 2009, he joined as a Faculty Member of the School of Electrical and Computer Engineering, NTUA, where he is currently a Full Professor. From 2004 to 2009, he was an Assistant Professor with the School of Production Engineering and Management, Technical University of Crete, Greece. From 1994 to 2003, he was with Schneider Electric AE, the Greek Subsidiary of Schneider Electric, where he was a Quality Control Engineer, a Transformer Design Engineer, the Research and Development Manager, and the Low Voltage Products Marketing Manager. His current research interests include optimization algorithms and computational intelligence techniques for the optimal operation and planning of smart distribution systems. He is a member of the Technical Chamber of Greece.
\end{IEEEbiography}

\begin{IEEEbiography}[{\includegraphics[width=1.1in,height=1.25in,clip,keepaspectratio]{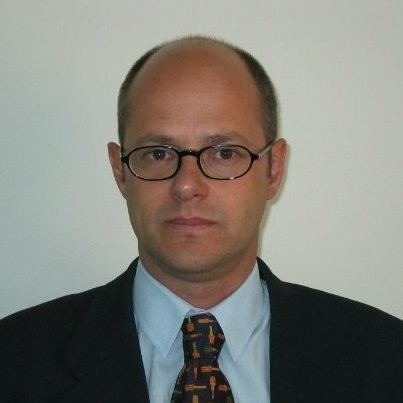}}]{DIMITRIS ASKOUNIS}  is currently a Professor
with the School of Electrical and Computer Engineering, National Technical University of Athens
(NTUA), and the Deputy Director of the Decision
Support Systems Laboratory. He has over 25 years
of experience in decision support systems, intelligent information systems and manufacturing,
e-business, e-government, open and linked data,
big data analytics, AI algorithms, and application
of modern IT techniques in the management of
companies and organizations. He was the Scientific Director of over 50 European research projects in the above areas (FP7, Horizon2020, etc.). He has
participated in many projects and activities in NIS and MEDA countries
related to the monitoring and evaluation of major projects, training of business executives, and development of IT systems. For a number of years,
he was the Advisor to the Minister of Justice and the Special Secretary for
Digital Convergence for the introduction of information and communication
technologies in public administration. Since June 2019, he has been the
President of the Information Society SA.
\end{IEEEbiography}


\EOD

\end{document}